\documentclass[10pt,twocolumn,letterpaper]{article}

\usepackage[pagenumbers]{iccv} 
\newcommand{\figref}[1]{Fig.~\ref{#1}}
\newcommand{\tabref}[1]{Tab.~\ref{#1}}
\newcommand{\secref}[1]{Section~\ref{#1}}
\renewcommand{\eqref}[1]{Eqn.~(\ref{#1})}

\definecolor{iccvblue}{rgb}{0.21,0.49,0.74}
\usepackage[pagebackref,breaklinks,colorlinks,allcolors=iccvblue]{hyperref}
\usepackage{multirow}
\usepackage{pifont}
\usepackage{adjustbox}
\usepackage{appendix}
\usepackage{algorithm}
\usepackage{algpseudocode}
\usepackage{amsmath} 
\usepackage{float}

\usepackage{float}
\usepackage{pifont}
\usepackage{array}
\usepackage{xcolor}
\usepackage{colortbl}
\usepackage[framemethod=TikZ]{mdframed}
\usepackage{etoolbox}
\usepackage[listings]{tcolorbox}
\tcbuselibrary{listings,theorems}
\definecolor{UserExampleBg}{HTML}{FAFDF9}
\newmdenv[
    roundcorner=5pt,
    backgroundcolor=UserExampleBg,
    linecolor=black,
    outerlinewidth=1pt,
    frametitlebackgroundcolor=black,
    frametitlefont={\bfseries\color{white}},
]{user_example}
\AtBeginEnvironment{user_example}{\setlength{\parindent}{0pt}}
\newtcbtheorem[]{prompt}{Prompt}%
{colback=green!5,colframe=black,fonttitle=\bfseries, left=.02in, right=.02in,bottom=.02in, top=.02in}{prompt}

\def\paperID{4795} % *** Enter the Paper ID here
\def\confName{ICCV}
\def\confYear{2025}

\title{LeX-Art: Rethinking Text Generation via Scalable High-Quality Data Synthesis}
\author{
Shitian Zhao\textsuperscript{1}\thanks{Equal Contribution} \quad
Qilong Wu\textsuperscript{1}$^\ast$ \quad
Xinyue Li\textsuperscript{1}$^\ast$ \quad
Bo Zhang\textsuperscript{1} \quad
Ming Li\textsuperscript{1} \quad
Qi Qin\textsuperscript{1} \quad
Dongyang Liu\textsuperscript{1,2} \\
Kaipeng Zhang\textsuperscript{1}  \quad
Hongsheng Li\textsuperscript{2}  \quad
Yu Qiao\textsuperscript{1}  \quad
Peng Gao\textsuperscript{1}  \quad
Bin Fu\textsuperscript{1}\thanks{Corresponding Authors}  \quad
Zhen Li\textsuperscript{2}$^\dagger$ \\
\textsuperscript{1}Shanghai AI Laboratory \quad
\textsuperscript{2}The Chinese University of Hong Kong \\
\url{https://zhaoshitian.github.io/lexart/}
}

\begin{document}
% \maketitle

\twocolumn[{
\renewcommand\twocolumn[1][]{#1}
\maketitle
\begin{center}
    \centering
    \vspace*{-.8cm}
    \includegraphics[width=\textwidth]{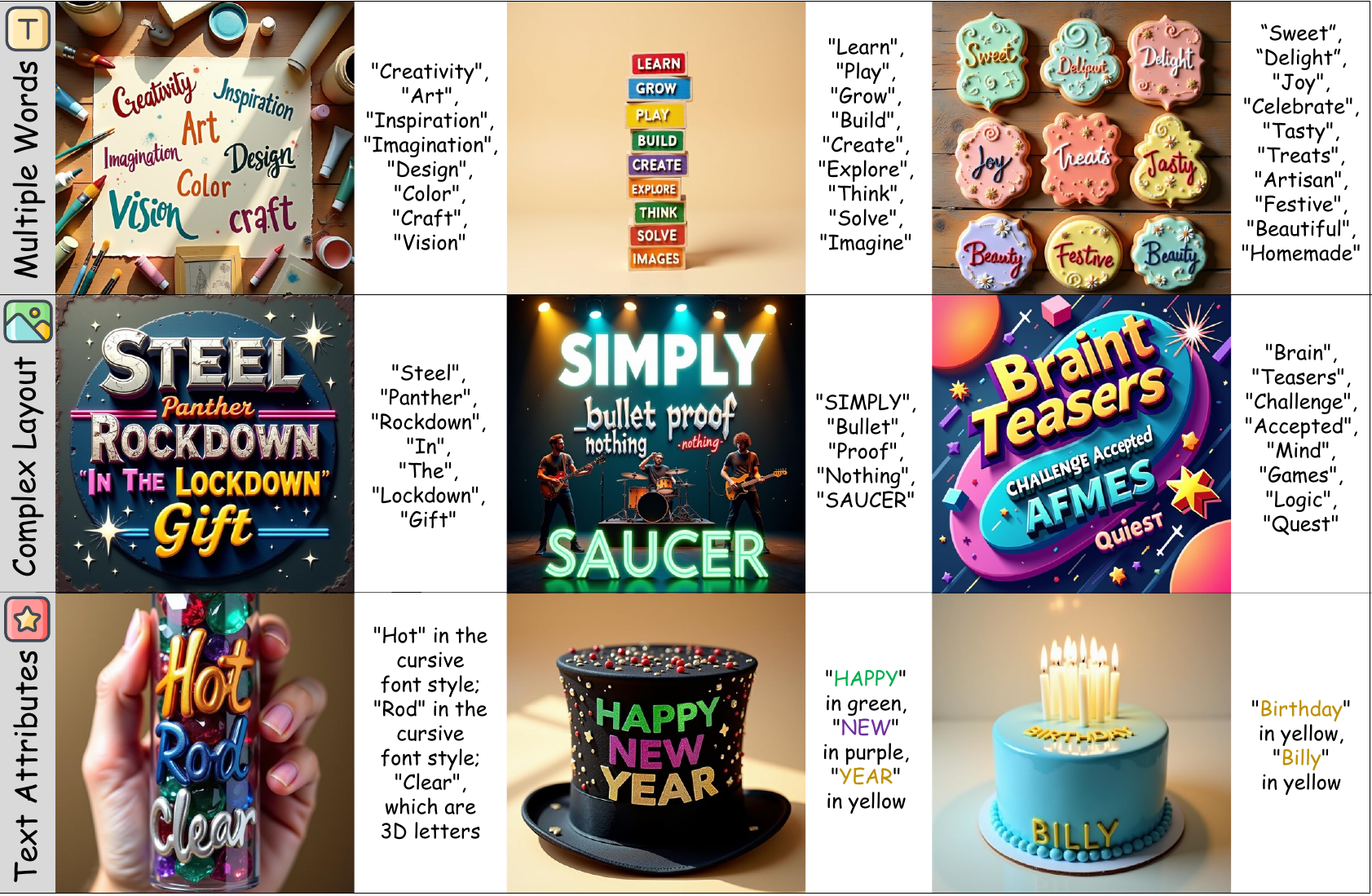}
    \vspace*{-.6cm}
    \captionof{figure}{Given the prompts for visual text generation, our proposed \textit{LeX-FLUX} and \textit{LeX-Lumina} can generate text images with \textbf{multiple words}, \textbf{aesthetic complex layout}, and \textbf{good text attributes controllability}.}
\label{fig:teaser}
\end{center}
}]
\let\thefootnote\relax\footnotetext{$^*$ Equal Contribution \hspace{3pt} $^\dagger$ Corresponding authors
}

% \begin{figure*}[h]
%     \centering
%     \includegraphics[width=\linewidth]{ICCV2025-Author-Kit-Feb/figures/demo2.pdf}
%     \caption{
%     % Showcase of text rendering results from LeX-Lumina (first two rows) and LeX-FLUX (last two rows) on text-to-image tasks. The examples demonstrate the models’ ability to generate clear, well-aligned, and aesthetically pleasing text within images. 
%     LeX-FLUX and LeX-Lumina have shown advanced text rendering capability in terms of multiple words, complex layout, and text attributes controllability.}
%     \label{fig:showcase}
% \end{figure*}

\begin{abstract}
% In this work, we try to address the challenging task of generating visual text images where the text is not only accurate but also aesthetically appealing. Existing methods struggled with this balance due to complex model architectures and reliance on large-scale but low-quality training data. To tackle these limitations, we first distill DeepSeek-R1 into a 14B-parameter prompt enhancer, \textbf{\textit{LeX-Enhancer}}, significantly improving structured textual descriptions for text-to-image generation. Second, we curate \textbf{\textit{LeX-10K}}, a systematically constructed, high-quality dataset focusing on enhancing text fidelity, aesthetic layout, and stylistic diversity. Third, leveraging \textit{LeX-10K}, we develop two specialized models, \textbf{\textit{LeX-Flux}} (12B) and \textbf{\textit{LeX-Lumina}} (2B), achieving state-of-the-art performance in visual text rendering. Fourth, we propose \textbf{\textit{LeX-Bench}}, a comprehensive benchmark along with Pairwise Normalized Edit Distance ( \textbf{\textit{PNED}}), a novel metric designed to evaluate text correctness and aesthetic consistency more effectively. Experiments demonstrate substantial performance gains: \textit{LeX-Lumina} improves PNED by 22.16\% on standard benchmarks, and \textit{LeX-Flux} achieves improvements of 10.32\% in color accuracy, 5.60\% in positional accuracy, and 5.63\% in font fidelity. Models and codes are available at \href{https://github.com/zhaoshitian/LeX-Art}{https://github.com/zhaoshitian/LeX-Art}.

We introduce LeX-Art, a comprehensive suite for high-quality text-image synthesis that systematically bridges the gap between prompt expressiveness and text rendering fidelity. Our approach follows a data-centric paradigm, constructing a high-quality data synthesis pipeline based on Deepseek-R1 to curate LeX-10K, a dataset of 10K high-resolution, aesthetically refined 1024$\times$1024 images. Beyond dataset construction, we develop LeX-Enhancer, a robust prompt enrichment model, and train two text-to-image models, LeX-FLUX and LeX-Lumina, achieving state-of-the-art text rendering performance. To systematically evaluate visual text generation, we introduce LeX-Bench, a benchmark that assesses fidelity, aesthetics, and alignment, complemented by Pairwise Normalized Edit Distance (PNED), a novel metric for robust text accuracy evaluation. Experiments demonstrate significant improvements, with LeX-Lumina achieving a 79.81\% PNED gain on CreateBench, and LeX-FLUX outperforming baselines in color (+3.18\%), positional (+4.45\%), and font accuracy (+3.81\%). Our codes, models, datasets, and demo are publicly available.
% Models and codes: \href{https://github.com/zhaoshitian/LeX-Art}{https://github.com/zhaoshitian/LeX-Art}.

% Contributions: \\
% 1)We find that DeepSeek-R1 significantly enhances prompt quality for text-to-image generation and distill it into a 14B parameter prompt enhancer.\\
% 2)we curate a high-quality, high-aesthetic dataset through a systematic data curation pipeline.\\
% 3)we develop Flux-Text and Lumina-Text, a series of end-to-end trained models that achieve state-of-the-art text rendering performance.\\
% 4)we introduce LeX-Bench and propose MNED, a new evaluation metric that better measures text correctness and aesthetic consistency.\\

\end{abstract}  
% \begin{figure*}[!t]
%     \centering
%     \includegraphics[width=0.95\linewidth]{ICCV2025-Author-Kit-Feb/figures/demo1.pdf}
%     \caption{
%     % Showcase of text rendering results from LeX-Lumina (first two rows) and LeX-FLUX (last two rows) on text-to-image tasks. The examples demonstrate the models’ ability to generate clear, well-aligned, and aesthetically pleasing text within images. 
%     LeX-FLUX and LeX-Lumina have shown advanced text rendering capability in terms of multiple words, complex layout, and text attributes controllability.}
%     \label{fig:showcase}
% \end{figure*}

% \begin{figure*}[!t]
%     \centering
%     \includegraphics[width=0.9\linewidth]{ICCV2025-Author-Kit-Feb/figures/demo1.pdf}
%     \caption{
%     % Showcase of text rendering results from LeX-Lumina (first two rows) and LeX-FLUX (last two rows) on text-to-image tasks. The examples demonstrate the models’ ability to generate clear, well-aligned, and aesthetically pleasing text within images. 
%     LeX-FLUX and LeX-Lumina have shown advanced text rendering capability in terms of multiple words, complex layout, and text attributes controllability.}
%     \label{fig:showcase}
% \end{figure*}

\section{Introduction}
\label{sec:intro}

Visual text generation is an important task for text-to-image (T2I) models. Producing text images that are clear, accurate, and aesthetically pleasing significantly enhances productivity in the design industry. Previous efforts, constrained by the limitations of foundational models such as SD-1.5~\cite{rombach2022high} and SD-XL~\cite{podell2023sdxl} have resulted in poor performance in visual text generation. Consequently, researchers has focused predominantly on improving the accuracy of text, while overlooking an equally important aspect: \textit{the aesthetic quality of the generated text and its seamless integration and interaction with image content.}

Due to the limitations of base T2I models, recent works have adopted control-based~\cite{zhang2023adding} approaches for accurate visual text rendering, incorporating glyph information into the generation pipeline through specialized glyph modules, \eg, AnyText~\cite{tuo2023anytext,tuo2024anytext2}, GlyphControl~\cite{yang2023glyphcontrol}, Text-Diffuser~\cite{chen2023textdiffuser,chen2024textdiffuser}, and Glyph-ByT5~\cite{liu2024glyphbyt5,liu2024glyphbyt5v2}.
Incorporating glyph information in the image generation process can enhance control over text accuracy and layout in generated images, which is particularly beneficial for paragraph text generation. 
However, these methods often compromise diversity, aesthetics, and seamless integration with surrounding visual content. While ensuring text accuracy—regardless of its fusion with the background—may suffice for structured text scenarios like poster design, it falls short in applications requiring minimal yet visually harmonious text, such as slogans, logos, and artistic typography. In these cases, aesthetic appeal, integration with the overall image, and dynamic layout become crucial.

To achieve \textit{high-quality multi-word visual text generation with diverse font styles, dynamic layouts, and strong aesthetic appeal}, we take an alternative approach that avoids the constraints of control-based methods. Instead of introducing additional control signals, we aim to maximize the inherent text rendering capabilities of base T2I models through high-quality data curation and supervised fine-tuning. 

Existing open-sourced and web-crawled text image datasets~\cite{gu2022wukong,tuo2023anytext,tuo2024anytext2,schuhmann2021laion} exhibit significant limitations in enhancing the text rendering capabilities of foundation models. This is primarily due to the scarcity of high-quality text image samples within these datasets, making it challenging to meet the demands of model training. Given these limitations, relying solely on curated datasets is insufficient for achieving high-quality text rendering. Instead, a more scalable and systematic approach is required to generate diverse, aesthetically rich, and well-integrated text samples. To address these challenges, we introduce \textit{\textbf{LeX-Art}}, a comprehensive framework for high-quality text-image synthesis and enhancement, designed to systematically bridge the gap between prompt expressiveness and high-quality rendered text.

As shown in \figref{fig:suite}, LeX-Art consists of three key components: (1) dataset construction, (2) model fine-tuning, and (3) benchmark evaluation. Our approach begins by leveraging  the strong reasoning capabilities of DeepSeek-R1~\cite{guo2025deepseek} to refine seed prompts extracted from the AnyWord-3M~\cite{tuo2023anytext} dataset. These enriched prompts incorporate detailed textual attributes such as font styles, color schemes, and spatial layouts, significantly improving the quality of textual descriptions for image synthesis.
With these enhanced prompts, we construct LeX-10K, a dataset of 10K high-quality text-image pairs. To ensure data reliability, we introduce a multi-stage filtering and recaptioning process: Q-Align~\cite{wu2023qalign} and Paddle-OCR-v3~\cite{li2022pp} assess fidelity, aesthetic quality, and text bounding box coverage, while a knowledge-augmented recaptioning module refines textual descriptions to enhance alignment with generated images.
Beyond dataset creation, we leverage 60,856 DeepSeek-R1-enhanced prompts to fine-tune a locally deployable model, LeX-Enhancer, which specializes in robust prompt enrichment for text-image generation. We further fine-tune two visual text generation models, LeX-FLUX and LeX-Lumina, using LeX-10K. Notably, our experiments demonstrate that LeX-FLUX achieves superior text rendering performance, while even the lightweight LeX-Lumina (2B) benefits significantly, highlighting the impact of our data synthesis strategy.
To systematically evaluate our approach, we propose LeX-Bench, a benchmark designed to assess visual text generation across multiple dimensions, including text fidelity, aesthetics, and alignment with input prompts.
Furthermore, we introduce Pairwise Normalized Edit Distance (PNED), a new metric for evaluating text accuracy. PNED computes the discrepancy between ground-truth and OCR-detected words using NED-based~\cite{yujian2007ned} matching, making it a flexible metric for non-glyph-conditioned visual text generation.
Experimental results demonstrate that LeX-Art provides a scalable and effective framework for improving text-image generation, offering a systematic solution to the challenges of high-quality text rendering.

\section{Related Work}
\label{sec:formatting}

\noindent
\textbf{Text-to-image models.}
% Image generation has always been an important task in the AI community. From DDPM~\cite{ho2020ddpm} to Flow Matching~\cite{lipman2022flow} to autoregressive model~\cite{razavi2019vqvae2, van2017vqvae,sun2024llamagen}, image generation modeling paradigm transferred progressively. Also, the model architecture was developed from U-Net~\cite{ronneberger2015unet} to DiT~\cite{peebles2023DiT}. The training datasets start from the web-crawled low-quality data, \eg, LAION-5B~\cite{schuhmann2021laion}, to high-quality human-curated datasets, \eg, JourneyDB~\cite{sun2023journeydb}. Advanced text-to-image generation models are running into our sight with better image generation ability, \eg, high quality, better controllability, higher aesthetic, \eg, FLUX.1 [dev]\cite{flux2024}, Lumina-Next~\cite{zhuo2024lumina}, Lumina-mGPT~\cite{liu2024lumina}, SANA~\cite{xie2024sanaefficienthighresolutionimage}, Janus-Pro~\cite{chen2025janusprounifiedmultimodalunderstanding}, Stable Diffusion 3~\cite{esser2024sd3}, Hunyuan-Dit~\cite{li2024hunyuan}, Kolors~\cite{kolors}, Seedream~\cite{gong2025seedream20}.\\ 
Image generation has long been a pivotal problem in the field of artificial intelligence. To efficiently and effectively generate high-quality images, researchers have developed a variety of algorithms. These include Generative Adversarial Networks (GANs)~\cite{goodfellow2020gan,zhu2017cyclegan,isola2017i2i,brock2018largegan,karras2019stylegan,wang2018esrgan}; Variational Autoencoders (VAEs)~\cite{an2015variational,kingma2013vae,kusner2017grammar,razavi2019vqvae2,van2017vqvae,vahdat2020nvae}; and Diffusion Models~\cite{ho2020ddpm,baldridge2024imagen,chen2025janusprounifiedmultimodalunderstanding,gong2025seedream20}. However, in recent years, Flow Matching~\cite{lipman2022flow,liu2022flow} has emerged as the de facto paradigm in image generation due to its training stability and sampling efficiency. 

% In addition to advances in modeling paradigms, the architectures used for image generation have also evolved significantly. Initially, U-Net~\cite{ronneberger2015unet}, with its ability to model pixel-to-pixel relationships, was widely adopted in image generation tasks. During the past two years, Sora~\cite{sora} demonstrated the scalability of the DiT~\cite{peebles2023DiT} architecture in both image and video generation, triggering a surge of research focused on image generation models based on the DiT framework. 

As described above, with continuous advancements in modeling paradigms and architectural designs, image generation models have achieved remarkable progress in generating realistic portraits, high-resolution images, and text image synthesis. Among these, text rendering has become increasingly important as a challenging task with significant economic value, making it a key metric for evaluating the capabilities of text-to-image models. 
In SD3~\cite{esser2024sd3} and Seedream~\cite{gong2025seedream20}, they use DPO~\cite{rafailov2023dpo,wallace2024diffusion} to improve the model's text rendering ability. In Playground-v3~\cite{liu2024playgroundv3}, they use a fine-grained caption to improve the text rendering capability. In this work, we focus on improving the text rendering capabilities of T2I models via post-training with high-quality synthetic data.

% In this work, we focus on improving the text rendering capabilities of T2I models, aiming to push the boundaries of performance in this critical dimension.

\noindent
\textbf{Visual text generation.}
% Besides fundamental image generation, visual text generation is always receiving much attention due to its huge commercialization potential and its difficulty. The main obstacle was considered, as the information of the text that needs to be rendered would be harmed after the text encoding. The researchers tried to alleviate this issue from different perspectives: (1) using a more powerful text encoder, \eg, T5~\cite{raffel2020exploring}, as the text encoding module in the image generation model. On this line, there is Imagen~\cite{baldridge2024imagen}; (2) another line of work brings the glyph information into the image generation pipeline, \eg, GlyohControl~\cite{yang2023glyphcontrol}, AnyText~\cite{tuo2023anytext,tuo2024anytext2}, and Text-Diffuser~\cite{chen2023textdiffuser,chen2024textdiffuser}. 
A prevailing explanation for the difficulty of text rendering in text-to-image generation models is that the text information in the prompt, after being processed by the text encoder, often becomes distorted or inadequately preserved. This degradation prevents the model from effectively rendering textual content within the generated image. To address this issue, much work has focused on architectural modifications~\cite{chen2023textdiffuser,chen2024textdiffuser,liu2024glyphbyt5,liu2024glyphbyt5v2,ma2023glyphdraw,tuo2023anytext,tuo2024anytext2,yang2023glyphcontrol}, such as incorporating additional modules to explicitly inject glyph information into the image generation process.

However, models like FLUX~\cite{flux2024,flux11pro} have demonstrated that even without glyph modules, a model conditioned solely on prompts can achieve decent text rendering capabilities. Furthermore, most contemporary text-to-image models already possess a baseline ability to render text~\cite{flux2024,esser2024sd3,flux11pro,gong2025seedream20,kolors,li2024hunyuan}, though they struggle with challenges such as multi-text scenarios, fine-grained text attribute control, and complex layouts. 

Given this, instead of introducing architectural modifications, we explore an alternative perspective by enhancing the model's text rendering capability through data-centric approaches. Specifically, we aim to improve the model's ability to handle the aforementioned challenges by carefully designing and augmenting the training data. This approach leverages the inherent potential of existing model architectures while addressing their limitations in text rendering.

\section{LeX-Art}

\begin{figure}[htbp]
    \centering
    \includegraphics[width=0.9\linewidth]{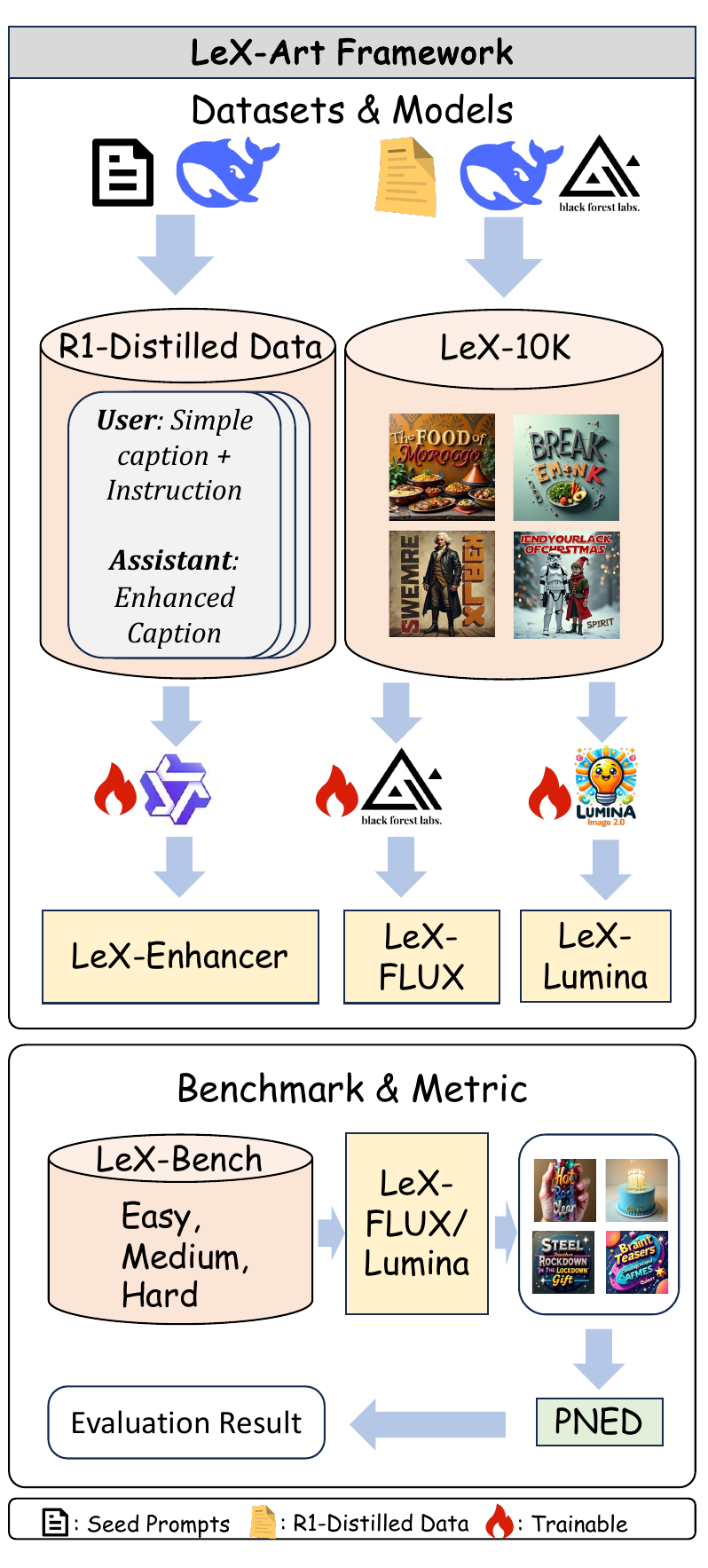}
    \caption{\textbf{Illustration of LeX-Art Framework.}}
    \label{fig:suite}
    \vspace{-1em}
\end{figure}

\begin{figure*}[thbp]
    \centering
    \includegraphics[width=0.9\linewidth]{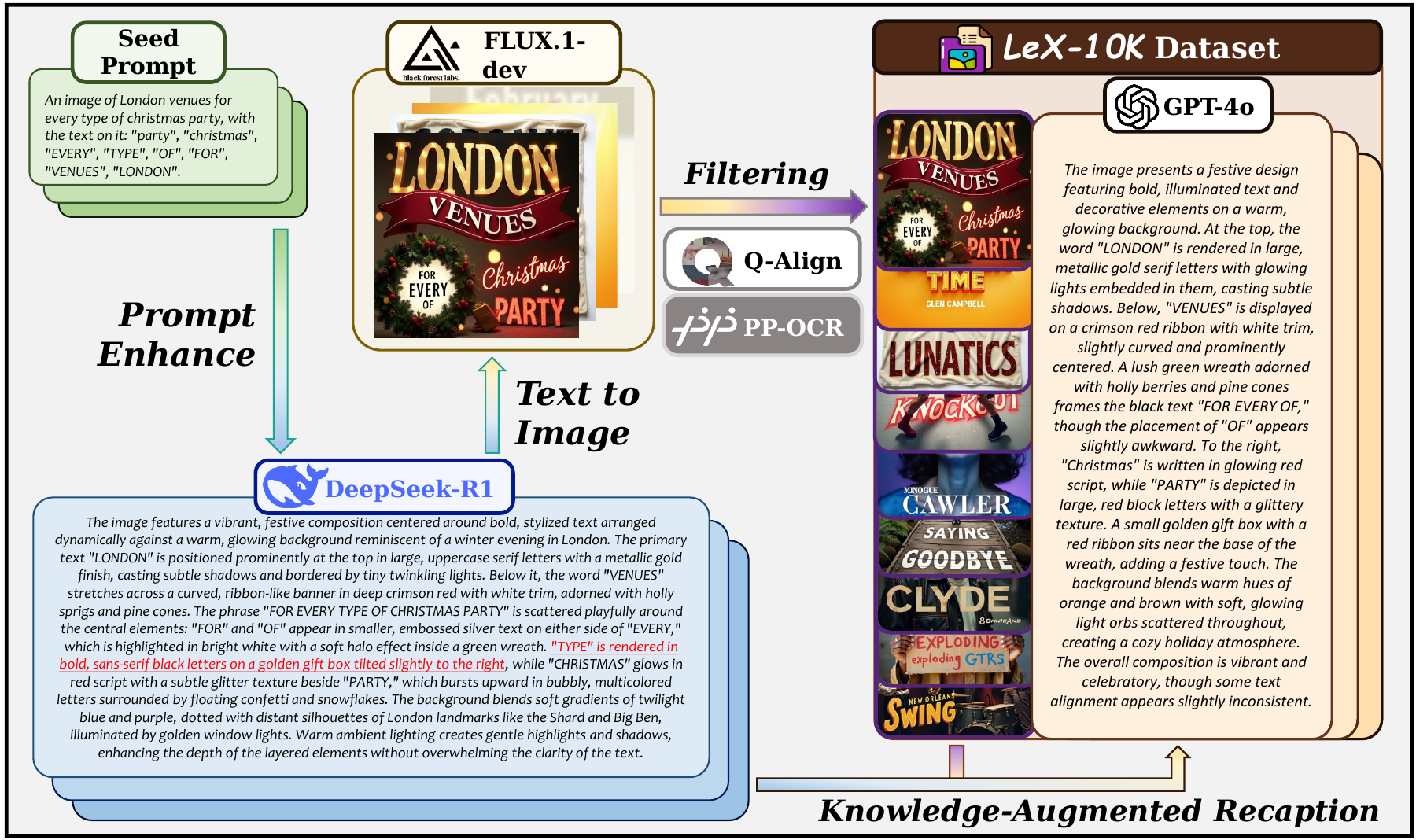}
    \caption{\textbf{The framework of data construction pipeline.} The \textcolor{red}{\underline{red}} words in the R1 enhanced prompt are not rendered in the generated image, and it is fixed after the knowledge-augmented recaption by gpt-4o.}
    \label{fig:overview}
\end{figure*}

\begin{figure*}[thbp]
    \centering
    \includegraphics[width=0.95\linewidth]{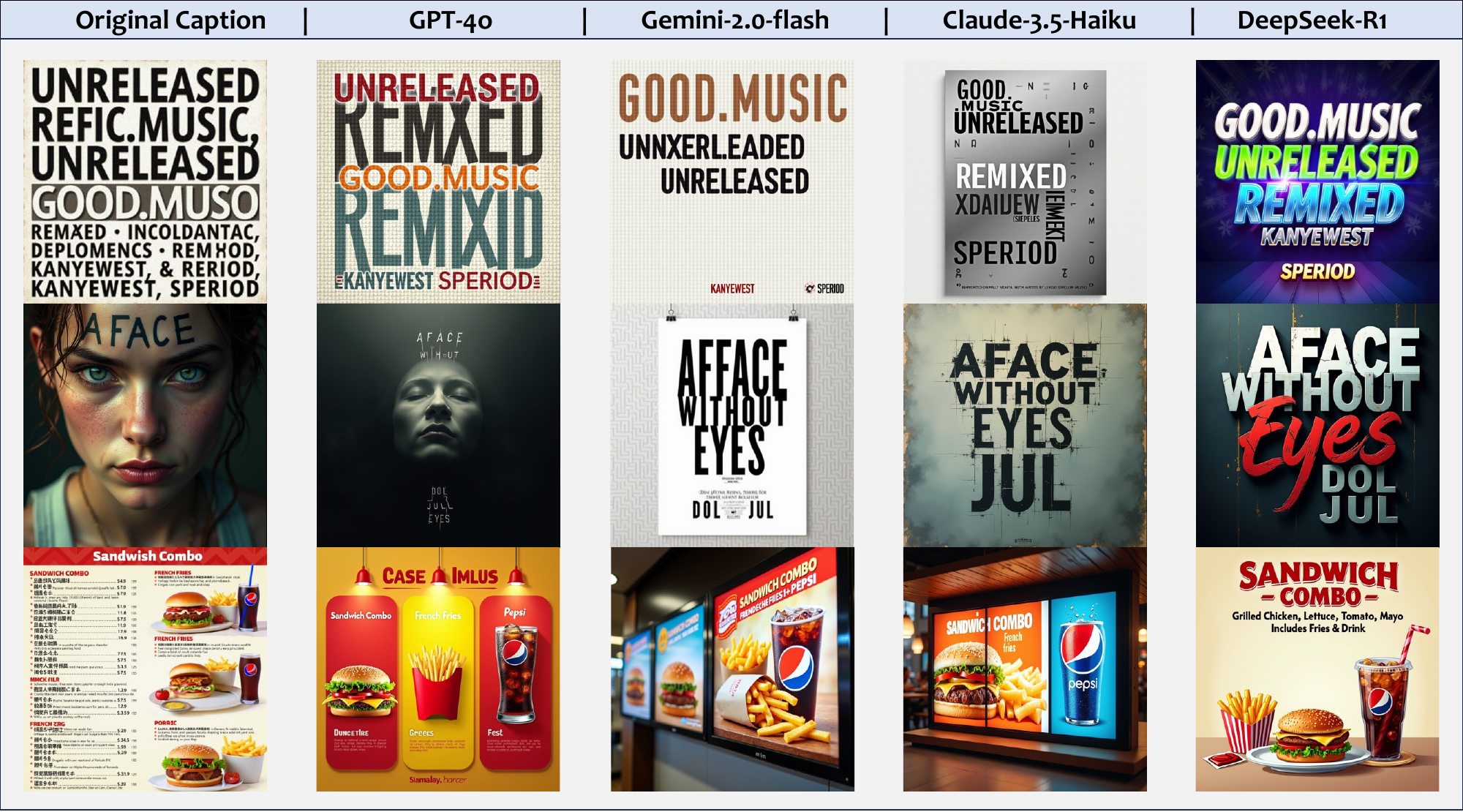}
    \caption{\textbf{Images generated by FLUX.1 [dev]~\cite{flux2024} based on different prompts.} The origin caption from the first raw to the bottom raw: (1) A poster with the words Good Music remixed and unreleased on it, with text on it: ``UNRELEASED", ``REMIXED", ``GOOD.MUSIC", ``KANYEWEST", ``SPERIOD". (2) A movie poster, with text on it: ``AFACE", ``WITHOUT", ``EYES", ``DOL", ``JUL". (3) A menu of a fast food restaurant that contains ``Sandwich Combo", ``Grilled Chicken", ``Lettuce", ``Tomato", ``Mayo", ``Fries\&Drink", and ``Pepsi".}
    \label{fig:r1}
\end{figure*}

\begin{figure}[thbp]
    \centering
    \includegraphics[width=0.95\linewidth]{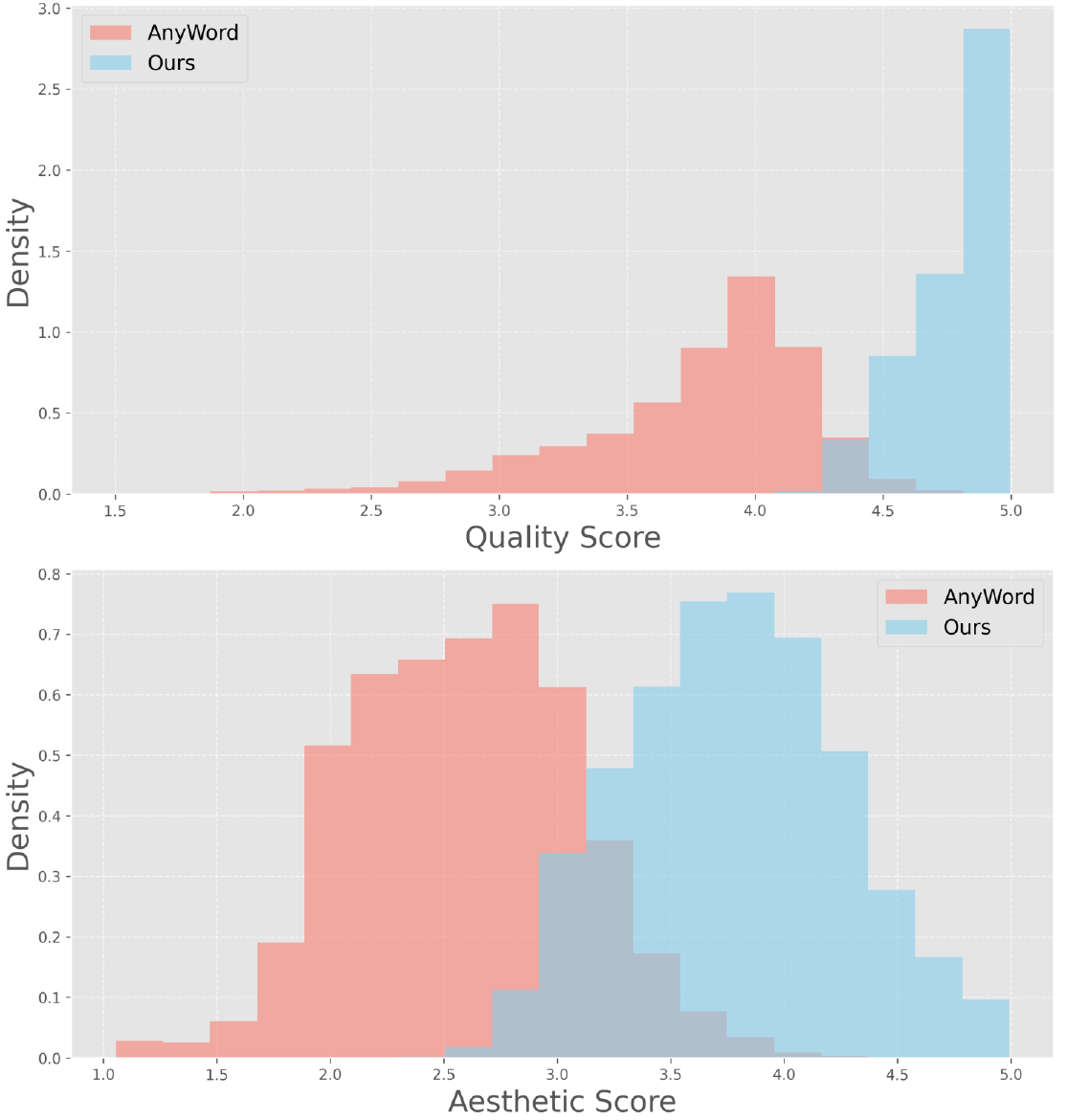}
    \caption{\textbf{Image quality score and image aesthetics score distribution of AnyText dataset~\cite{tuo2023anytext} and LeX-10K.} We randomly sampled 10K data entries from AnyWord-3M. Using Q-Align~\cite{wu2023qalign}, we calculated the quality scores and aesthetic scores for these 10K data entries along with the images in LeX-10K, and visualized the distributions of these two types of scores. We observed that LeX-10K generally has higher quality scores and aesthetic scores overall.}
    \label{fig:statistics}
\end{figure}

To tackle the challenges of accurate and aesthetic text rendering in image generation, we propose \textbf{LeX-Art}, a unified framework that enhances both data quality and model capability. LeX-Art consists of four components: (a) \textbf{LeX-10K}, a high-quality dataset built through prompt refinement and multi-stage filtering; (b) \textbf{LeX-Enhancer}, a prompt enrichment model, along with two fine-tuned generation models, \textbf{LeX-FLUX} and \textbf{LeX-Lumina}; (c) \textbf{LeX-Bench}, a benchmark for evaluating fidelity, aesthetics, and alignment; and (d) \textbf{PNED}, a new metric for flexible text accuracy evaluation. Together, as shown in \figref{fig:suite}, these modules form a scalable solution for improving visual text generation.

% To address the longstanding challenges in high-quality text-image generation—such as inaccurate text rendering, limited prompt diversity, and lack of standardized evaluation—we introduce LeX-Art, a unified framework that integrates data, models, and benchmarks to systematically enhance visual text generation. At the core of our approach is the insight that high-quality, attribute-rich prompts can significantly boost the performance of text-to-image models. Building on this, LeX-Art consists of four key components: (a) LeX-10K, a curated dataset of 10,000 high-quality text-image pairs created via prompt refinement and multi-stage filtering; (b) LeX-Enhancer, a prompt enrichment model trained on 60K DeepSeek-R1-refined samples, along with two fine-tuned visual text generation models, LeX-FLUX and LeX-Lumina; (c) LeX-Bench, a benchmark designed to evaluate fidelity, aesthetic quality, and alignment; and (d) a novel metric, Pairwise Normalized Edit Distance (PNED), tailored to quantify text accuracy in a flexible and OCR-compatible manner. Together, as shown in \figref{fig:suite}, these components form a scalable pipeline for improving the quality and controllability of text-image generation.
% LeX-Art is a comprehensive suite comprising: a) LeX-10K Dataset, b) LeX-Enhancer, LeX-FLUX and LeX-Lumina, c) LeX-Bench, and d) Pairwise Normalized Edit Distance (PNED) metric, as below.

\subsection{LeX-10K dataset}
\label{sec:dataset}

Existing web-crawled text image datasets, such as AnyWord-3M~\cite{tuo2023anytext,tuo2024anytext2} and MARIO-10M~\cite{chen2023textdiffuser,chen2024textdiffuser}, suffer from several key limitations, including inconsistent and low resolution (typically below $1024 \times 1024$), frequent blurriness from direct online sourcing, lack of aesthetic curation resulting in unbalanced and unappealing layouts, and shallow captions that overlook crucial visual details like font, color, and text positioning. To address these issues, we introduce LeX-10K, a high-quality text image dataset constructed through a structured four-stage pipeline: (1) enhancing prompts using DeepSeek-R1~\cite{guo2025deepseek}, (2) filtering and selecting high-resolution, visually pleasing images, (3) refining captions via knowledge-augmented recaptioning with GPT-4o~\cite{hurst2024gpt}. The data construction pipeline is shown in \figref{fig:overview}.

\paragraph{DeepSeek-R1 as a prompt enhancer.}
In text-to-image generation, enriching prompts with fine-grained visual details has been shown to significantly improve output quality~\cite{ideogram}. Motivated by this, we first curated a large-scale dataset of millions of simple captions extracted from text images. To ensure data quality, we filtered out OCR results that were empty, meaningless (\eg, ``*"), or overly verbose (\eg, exceeding 50 words), and retained only captions with fewer than 15 words.

To enrich these simple captions with detailed visual attributes, we prompted various large language models (LLMs) using a well-designed instruction (see Appendix~\ref{appendix:instruction}). As shown in \figref{fig:r1}, we evaluated GPT-4o~\cite{hurst2024gpt}, Gemini-2.0-flash~\cite{gemini2}, Claude3.5-Haiku~\cite{claude35}, and DeepSeek-R1~\cite{guo2025deepseek}. Among them, DeepSeek-R1 stood out by generating more fine-grained descriptions, including color, spatial layout, and font style of each text element. In contrast, other models tended to provide only high-level summaries of the text content. These detailed prompts are crucial for high-quality text image generation, underscoring the need for a rigorous selection process to retain the most effective outputs.

\paragraph{Synthetic image filtering.}
To mitigate the impact of random seeds on image quality, we generate five distinct images for each enhanced prompt using different seeds. Following the Best-of-N approach~\cite{wang2022self}, we first select the image with the highest quality and aesthetic scores. We then further discard any images in which the rendered text is too small to read.

To address issues of blurriness and poor readability observed in some generated images, we adopt a two-stage filtering strategy. First, we use Q-Align~\cite{wu2023qalign} to score each image based on a weighted combination of quality and aesthetic scores, with quality assigned twice the weight to emphasize clarity. The five images are then ranked by their overall scores, and the highest-scoring image is retained.
Second, since small rendered text correlates with lower visual clarity, we apply Paddle-OCR-v3~\cite{li2022pp} to detect all text regions and calculate their bounding box areas. We discard images where the largest text region is smaller than a threshold of 4000 pixels. After applying both filtering steps, we curate a final dataset of 10,000 high-quality images.

\paragraph{Knowledge-augmented recaptioning.}
While filtering improves image quality, ensuring accurate alignment between images and captions remains crucial. However, due to the limitations of FLUX.1~[dev]~\cite{flux2024}, the generated images do not always perfectly match the enhanced prompts—common issues include missing text, incorrect font styles, or wrong colors. To further improve the quality of image-text pairs, we leverage GPT-4o to refine the captions based on the actual visual content. Specifically, each image and its original caption are fed into GPT-4o, which is instructed to revise the caption to better reflect the image while preserving as much of the original content as possible. The detailed prompt instructions are provided in Appendix~\ref{appendix:postalign}. We find that this knowledge-augmented recaptioning process better preserves key visual details—such as font style and text attributes.

\paragraph{Data statistics.}
Following the generation, filtering, and knowledge-augmented recaptioning steps, we construct LeX-10K, a dataset comprising 10K high-quality text images paired with well-aligned, knowledge-rich captions. As illustrated in \figref{fig:statistics}, LeX-10K achieves significantly higher image quality and aesthetic scores compared to a randomly sampled subset of AnyWord-3M~\cite{tuo2023anytext}, highlighting its advantage in text rendering tasks.

\subsection{LeX-Enhancer, LeX-FLUX and LeX-Lumina}
To achieve high-quality text rendering in text-to-image (T2I) generation, both knowledge-rich prompts and strong T2I models are essential. To this end, we develop a prompt enhancer, LeX-Enhancer, and two optimized T2I models, LeX-FLUX and LeX-Lumina.

\textbf{LeX-Enhancer} is a lightweight prompt enhancement model distilled from DeepSeek-R1. Specifically, we collect 60,856 prompt pairs before and after R1 enhancement, termed as LeX-R1-60K, and fine-tune a Qwen2.5-14B model using LoRA~\cite{hu2022lora} to replicate the detailed prompting capabilities of R1. This enables efficient, large-scale generation of high-quality, visually grounded prompts.

To fully leverage the enhanced prompts, we fine-tune two strong T2I models on our curated LeX-10K dataset. \textbf{LeX-FLUX} is based on FLUX.1 [dev] (12B)~\cite{flux2024}, fine-tuned with a small batch size and conservative learning rate to ensure stable, high-fidelity generation. \textbf{LeX-Lumina}, built on Lumina-Image 2.0 (2B)~\cite{lumina2}, is optimized for lightweight deployment using a larger batch size and dropout strategy. Together, LeX-FLUX and LeX-Lumina offer a flexible balance between quality and efficiency for diverse deployment scenarios. Training details are provided in \textbf{Sec.}~\ref{sec:setting}.

\subsection{Pairwise Normalized Edit Distance (PNED)}
\label{subsub:pned}

While OCR Recall reflects whether the generated image contains text, it does not sufficiently capture the textual accuracy of Text-to-Image (T2I) models. To address this limitation, we introduce Pairwise Normalized Edit Distance (PNED)—a more flexible metric designed to assess how well the generated text matches the input prompt.

A commonly used metric, Normalized Edit Distance (NED) (Algorithm~\ref{alg:ned} in Appendix~\ref{appendix:NED}), is effective for evaluating text accuracy in glyph-conditioned models, where the generated text typically exhibits a strict one-to-one correspondence with the prompt. However, this assumption does not hold in non-glyph-conditioned models, where the text may appear in different orders or spatial arrangements.

To accommodate such variations, PNED treats both the prompt text and the OCR-extracted text as unordered sets of words rather than fixed sequences. It computes pairwise edit distances between words using the Hungarian Algorithm~\cite{hugarian}, then aggregates the matched word scores and applies penalties for unmatched words. The full procedure is described in Algorithm~\ref{alg:pned}. By leveraging PNED, we provide a more robust and generalizable metric for evaluating text accuracy across diverse T2I models.
% \subsection{Pairwise Normalized Edit Distance (PNED)}
% \label{subsub:pned}

% While OCR Recall measures the presence of generated text, it does not fully capture textual accuracy in T2I models. To address this, we introduce \textbf{Pairwise Normalized Edit Distance (PNED)}, a more flexible metric designed to evaluate how accurately the generated text aligns with the prompt.  

% A widely used metric, Normalized Edit Distance (NED) (Algorithm~\ref{alg:ned} in Appendix~\ref{appendix:NED}), is effective for assessing text accuracy in glyph-conditioned models, where the text in the generated image follows a strict one-to-one mapping with the prompt. However, this strict mapping assumption does not hold for non-glyph-conditioned models, where text order and placement can vary. 

% To provide a more adaptable evaluation, PNED considers the text in both the prompt and the OCR-extracted result as unordered lists rather than fixed sequences. It computes pairwise edit distances between words using the Hungarian Algorithm~\cite{hugarian}, summing the NED scores of matched words while penalizing unmatched words. The pseudo-algorithm for PNED is presented in Algorithm~\ref{alg:pned}. By leveraging PNED, we ensure a more generalizable and robust metric for evaluating text accuracy in diverse T2I models. 
\begin{algorithm}
\caption{Pairwise Normalized Edit Distance (PNED)}
\label{alg:pned}
\begin{algorithmic}[1]
\State \textbf{Input:} Prompt word set $X$ of size $n$, OCR result word set $Y$ of size $m$
\State Initialize cost matrix $C \in \mathbb{R}^{n \times m}$, with unmatched cost penalty defined as 1
\For{$i = 1$ to $n$}
    \For{$j = 1$ to $m$}
        \State $C_{i,j} \gets \text{NED}(X_i, Y_j)$
    \EndFor
\EndFor
\State $(r, c) \gets \text{HungarianAlgorithm}(C)$ \Comment{Find optimal word assignment}
\State $M \gets \sum_{k} C_{r_k, c_k}$ \Comment{Total cost of matched word pairs}
\State $U \gets |n - m| \times 1$ \Comment{Penalty for unmatched words}
\State \textbf{return} $(M + U)$ \Comment{Final score}
\end{algorithmic}
\end{algorithm}

\subsection{LeX-Bench}
\paragraph{Benchmark construction.}

To assess text rendering performance in text-to-image generation, we introduce \textbf{LeX-Bench}, a benchmark comprising 1,310 carefully designed prompts.
Each prompt contains two parts: an \textit{Image Caption} describing the image content, and a \textit{Text Caption} specifying the text to be rendered. The combined format is:
\textit{\{Image Caption\}, with the text on it: \{Text Caption\}.}, \eg, \textit{A picture of a blue and green abstract people logo on a purple background, with the text on it: ``AREA", ``PEOPLE".}  

	The prompts are derived from the AnyWord-3M~\cite{tuo2023anytext} dataset. The \textit{Image Caption} originates from the original image captions annotated by Florence-2~\cite{xiao2024florence}, and the \textit{Text Caption} is extracted using Paddle-OCR-v3~\cite{li2022pp}.
    % \footnote{When constructing LeX-10K, the seed prompts are sampled from AnyWord-3M~\cite{tuo2023anytext}. Here the samples used to build LeX-Bench are from the rest part of AnyWord-3M.}.
The samples used in LeX-Bench are non-overlapping with those used for LeX-10K.
	To ensure fluency, quality, and diversity, we filter out captions containing meaningless single letters or misspelled words, and then further refine the remaining prompts using GPT-4o for compatibility with T2I models. The exact instructions used in GPT-4o are provided in Appendix~\ref{appendix:lexbench}.
Finally, as shown in \figref{fig:lexbench}, LeX-Bench includes 1,310 prompts categorized by complexity: 630 Easy-Level (2–4 words), 480 Medium-Level (5–9 words), and 200 Hard-Level (10–14 words). The Easy-Level prompts also specify additional constraints such as text color, font, or position, detailed in Appendix~\ref{appendix:text_conditions}.

\begin{figure}
    \centering
    \includegraphics[width=\linewidth]{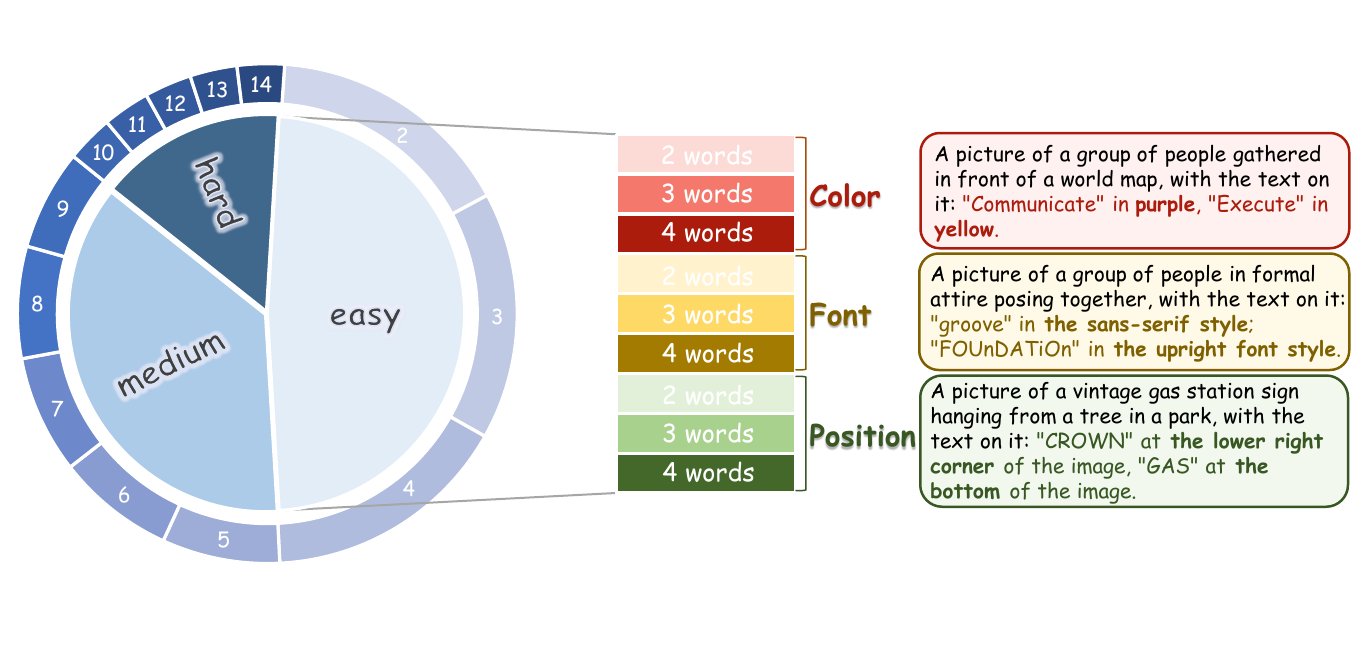}
    \caption{\textbf{Overview of LeX-Bench.} Prompts in LeX-Bench are split into three levels: 630 Easy-Level (2–4 words), 480 Medium-Level (5–9 words), and 200 Hard-Level (10–14 words). Prompts of the easy level also contain text attributes: color, font, position.}
    \label{fig:lexbench}
\end{figure}
    
\paragraph{Evaluation formulation.}
We evaluate model performance through three complementary methods: OCR-based evaluation, GPT-4o-based Visual Question Answering (VQA), and human preference assessment.
The OCR-based evaluation includes OCR Recall, which measures the match rate between generated and reference text, and our proposed Pairwise Normalized Edit Distance (PNED)~(\secref{subsub:pned}), designed to assess generation accuracy in non-glyph-conditioned T2I models.
For VQA-based evaluation, we use GPT-4o to assess whether the generated image satisfies specified attributes. This yields the VQA Score, with sub-metrics including Color-Score, Position-Score, and Font-Score.
Finally, we report the Preference Win Rate, which reflects human judgments on the quality and realism of the generated text images.
Detailed procedures for all evaluations are provided in Appendix~\ref{appendix: condition_eval}.

\section{Experiments}

\begin{table}[t]
    \centering
    \adjustbox{max width=0.45\textwidth}{
    \begin{tabular}{l|c|c|c}
    \toprule
       Methods & LeX-Enhancer& Acc.~$\uparrow$ & CLIPScore~$\uparrow$ \\
    \midrule
    ControlNet~\cite{yang2023glyphcontrol} &-&0.5837&0.8448 \\
    TextDiffuser~\cite{chen2023textdiffuser} &-&0.5921&0.8685 \\
    GlyphControl~\cite{yang2023glyphcontrol} &-&0.3710&0.8847 \\
       AnyText~\cite{tuo2023anytext}  &- &0.7239&0.8841 \\
    \hline
       \multirow{2}{*}{LeX-Lumina} & \ding{55}&0.3840&0.8100  \\
        & \ding{51}&0.6220&0.8832 \\
    \hline
       \multirow{2}{*}{LeX-FLUX}  & \ding{55}&0.5220&0.8754 \\
        & \ding{51}&0.7110&0.8918 \\
    \bottomrule
    \end{tabular}}
    \caption{\textbf{Comparison with glyph-conditioned models on AnyText-Benchmark~\cite{tuo2023anytext}.} We compare LeX-FLUX, LeX-Lumina and glyph-conditioned methods, \ie, ControlNet~\cite{yang2023glyphcontrol}, TextDiffuser~\cite{chen2023textdiffuser}, GlyphControl~\cite{yang2023glyphcontrol}, and AnyText~\cite{tuo2023anytext} on AnyText-Benchmark. We observe that glyph-conditioned methods perform better in terms of text rendering accuracy, due to the incorporation of glyph information. But even without the glyph information, our methods have a compatible performance with glyph-controlled methods and have a better performance on prompt-image alignment.}
    \label{tab:compare}
    % \vspace{-1em}
\end{table}

\begin{figure}[tbp]
    \centering
    \includegraphics[width=\linewidth]{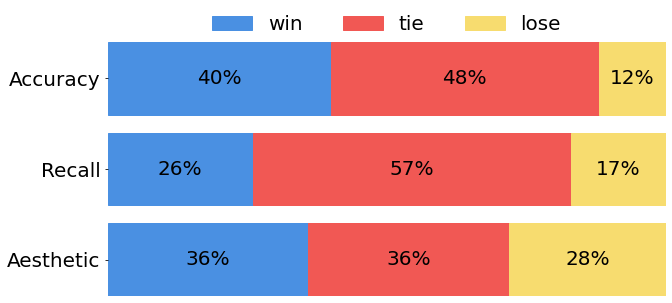}
    \caption{\textbf{Human preference result on text accuracy, text recall rate and aesthetics for LeX-Lumina.} For ease
 of illustration, we visualize the proportion of votes where LeX-Lumina wins, loses and ties with Lumina-Image 2.0.}
    \label{fig:human_preference}
\end{figure}

\subsection{Settings}
\label{sec:setting}

\paragraph{Implementation details.}
We curate 10K high-resolution (1024 $\times$ 1024) text-image dataset to finetune the FLUX.1 [dev] (12B)~\cite{flux2024} and Lumina-Image 2.0 (2B)~\cite{lumina2}.
All models are trained on 8 A100 GPUs using PyTorch’s Fully Sharded Data Parallel (FSDP) framework with $\mathrm{bf16}$ precision. LeX-FLUX is finetuned with a global batch size of 8, a learning rate of 1e-6, classifier-free guidance (CFG) scale of 1, and 6,000 training steps, while LeX-Lumina is trained with a global batch size of 256, a learning rate of 1e-4, CFG drop rate of 0.1, and 10,000 training steps. The prompt enhancer adopts LoRA with $\mathrm{rank}=64$ and $\alpha=32$, trained for 500 steps with a global batch size of 16, a learning rate of 1e-4, and a maximum prompt length of 5,120 tokens.
To support long sequences under limited resources, we enable sequence parallelism with a factor of 2.

\paragraph{Benchmarks and metrics.}
We evaluate our models, LeX-FLUX and LeX-Lumina, on several widely-used text rendering benchmarks, including SimpleBench~\cite{yang2023glyphcontrol}, CreateBench~\cite{yang2023glyphcontrol}, and AnyText-Benchmark~\cite{tuo2023anytext}. These benchmarks provide diverse text prompts and assess models using OCR-based metrics such as sentence accuracy (\textit{Sen. Acc}), normalized edit distance (NED)~\cite{yujian2007ned}, and OCR-F1. In addition, they report CLIP Score~\cite{hessel2021clipscore} and Frechet Inception Distance (FID)~\cite{heusel2017gans} to evaluate visual-textual alignment and image quality.
\begin{table*}[h]
    \centering
    \definecolor{deepgreen}{rgb}{0.0, 0.5, 0.0}
    \adjustbox{max width=0.95\textwidth}{
    \begin{tabular}{l|c|>{\centering\arraybackslash}m{1.5cm}>{\centering\arraybackslash}m{1.5cm}|>{\centering\arraybackslash}m{1.5cm}>{\centering\arraybackslash}m{1.5cm}||>{\centering\arraybackslash}m{1.5cm}>{\centering\arraybackslash}m{1.5cm}|>{\centering\arraybackslash}m{1.5cm}>{\centering\arraybackslash}m{1.5cm}}
        \toprule
        Datasets & Metric & \multicolumn{2}{c|}{FLUX.1 [dev]} & \multicolumn{2}{c||}{LeX-FLUX} & \multicolumn{2}{c|}{Lumina-Image-2.0} & \multicolumn{2}{c}{LeX-Lumina} \\
        \midrule
        LeX-Enhancer & - & \ding{55} & \ding{51} & \ding{55} & \ding{51} & \ding{55} & \ding{51} & \ding{55} & \ding{51} \\
        \midrule
        \multicolumn{10}{c}{\textit{Traditional Benchmark}} \\
        \midrule
        \multirow{3}{*}{SimpleBench} & PNED~$\downarrow$ & \cellcolor{deepgreen!54!white}0.74 & \cellcolor{deepgreen!89!white}0.37 & \cellcolor{deepgreen!4!white}1.30 & \cellcolor{deepgreen!90!white}0.35 & \cellcolor{deepgreen!75!white}0.45 & \cellcolor{deepgreen!79!white}0.47 & \cellcolor{deepgreen!0!white}2.24 & \cellcolor{deepgreen!78!white}0.51 \\
        & Recall~$\uparrow$ & \cellcolor{deepgreen!82!white}0.89 & \cellcolor{deepgreen!93!white}0.92 & \cellcolor{deepgreen!65!white}0.83 & \cellcolor{deepgreen!100!white}0.94 & \cellcolor{deepgreen!17!white}0.70 & \cellcolor{deepgreen!65!white}0.84 & \cellcolor{deepgreen!0!white}0.65 & \cellcolor{deepgreen!79!white}0.88 \\
        & Aesthetic~$\uparrow$ & \cellcolor{deepgreen!65!white}3.25 & \cellcolor{deepgreen!52!white}3.10 & \cellcolor{deepgreen!74!white}3.35 & \cellcolor{deepgreen!60!white}3.19 & \cellcolor{deepgreen!0!white}2.46 & \cellcolor{deepgreen!35!white}2.89 & \cellcolor{deepgreen!54!white}3.12 & \cellcolor{deepgreen!58!white}3.17 \\
        \hline
        \multirow{3}{*}{CreateBench} & PNED~$\downarrow$ & \cellcolor{deepgreen!32!white}2.60 & \cellcolor{deepgreen!62!white}1.68 & \cellcolor{deepgreen!41!white}2.36 & \cellcolor{deepgreen!65!white}1.45 & \cellcolor{deepgreen!0!white}5.40 & \cellcolor{deepgreen!38!white}2.76 & \cellcolor{deepgreen!37!white}1.98 & \cellcolor{deepgreen!70!white}1.09 \\
        & Recall~$\uparrow$ & \cellcolor{deepgreen!43!white}0.63 & \cellcolor{deepgreen!85!white}0.78 & \cellcolor{deepgreen!53!white}0.66 & \cellcolor{deepgreen!91!white}0.80 & \cellcolor{deepgreen!0!white}0.50 & \cellcolor{deepgreen!63!white}0.70 & \cellcolor{deepgreen!19!white}0.56 & \cellcolor{deepgreen!75!white}0.74 \\
        & Aesthetic~$\uparrow$ & \cellcolor{deepgreen!85!white}3.67 & \cellcolor{deepgreen!89!white}3.70 & \cellcolor{deepgreen!96!white}3.74 & \cellcolor{deepgreen!93!white}3.72 & \cellcolor{deepgreen!0!white}3.21 & \cellcolor{deepgreen!75!white}3.62 & \cellcolor{deepgreen!82!white}3.66 & \cellcolor{deepgreen!100!white}3.76 \\
        \hline
        \multirow{3}{*}{AnyText-1K\textsubscript{en}} & PNED~$\downarrow$ & \cellcolor{deepgreen!4!white}6.22 & \cellcolor{deepgreen!70!white}3.70 & \cellcolor{deepgreen!0!white}6.25 & \cellcolor{deepgreen!60!white}3.87 & \cellcolor{deepgreen!21!white}5.43 & \cellcolor{deepgreen!56!white}3.86 & \cellcolor{deepgreen!62!white}3.66 & \cellcolor{deepgreen!79!white}3.38 \\
        & Recall~$\uparrow$ & \cellcolor{deepgreen!20!white}0.51 & \cellcolor{deepgreen!88!white}0.70 & \cellcolor{deepgreen!26!white}0.52 & \cellcolor{deepgreen!90!white}0.71 & \cellcolor{deepgreen!0!white}0.35 & \cellcolor{deepgreen!70!white}0.60 & \cellcolor{deepgreen!13!white}0.38 & \cellcolor{deepgreen!83!white}0.62 \\
        & Aesthetic~$\uparrow$ & \cellcolor{deepgreen!48!white}3.23 & \cellcolor{deepgreen!90!white}3.50 & \cellcolor{deepgreen!59!white}3.30 & \cellcolor{deepgreen!90!white}3.50 & \cellcolor{deepgreen!0!white}2.92 & \cellcolor{deepgreen!69!white}3.36 & \cellcolor{deepgreen!83!white}3.45 & \cellcolor{deepgreen!100!white}3.56 \\
        \midrule
        \multicolumn{10}{c}{\textit{Our Proposed Benchmark}} \\
        \midrule
        \multirow{6}{*}{LeX-Bench\textsubscript{easy}} & Color~$\uparrow$ & \cellcolor{deepgreen!24!white}42.69 & \cellcolor{deepgreen!21!white}41.58 & \cellcolor{deepgreen!24!white}42.54 & \cellcolor{deepgreen!40!white}45.87 & \cellcolor{deepgreen!44!white}46.67 & \cellcolor{deepgreen!86!white}55.08 & \cellcolor{deepgreen!83!white}54.60 & \cellcolor{deepgreen!100!white}57.14 \\
        & Position~$\uparrow$ & \cellcolor{deepgreen!18!white}28.57 & \cellcolor{deepgreen!37!white}31.27 & \cellcolor{deepgreen!16!white}28.25 & \cellcolor{deepgreen!49!white}33.02 & \cellcolor{deepgreen!42!white}32.06 & \cellcolor{deepgreen!100!white}40.48 & \cellcolor{deepgreen!0!white}25.87 & \cellcolor{deepgreen!73!white}36.51 \\
        & Font~$\uparrow$ & \cellcolor{deepgreen!30!white}49.36 & \cellcolor{deepgreen!82!white}67.62 & \cellcolor{deepgreen!40!white}52.06 & \cellcolor{deepgreen!100!white}71.43 & \cellcolor{deepgreen!36!white}50.79 & \cellcolor{deepgreen!77!white}65.56 & \cellcolor{deepgreen!45!white}53.33 & \cellcolor{deepgreen!87!white}68.41 \\
        & PNED~$\downarrow$ & \cellcolor{deepgreen!31!white}1.71 & \cellcolor{deepgreen!67!white}1.16 & \cellcolor{deepgreen!25!white}1.73 & \cellcolor{deepgreen!75!white}1.06 & \cellcolor{deepgreen!0!white}1.83 & \cellcolor{deepgreen!55!white}1.25 & \cellcolor{deepgreen!33!white}1.54 & \cellcolor{deepgreen!80!white}1.01 \\
        & Recall~$\uparrow$ & \cellcolor{deepgreen!59!white}0.66 & \cellcolor{deepgreen!85!white}0.76 & \cellcolor{deepgreen!56!white}0.64 & \cellcolor{deepgreen!87!white}0.77 & \cellcolor{deepgreen!37!white}0.47 & \cellcolor{deepgreen!61!white}0.64 & \cellcolor{deepgreen!48!white}0.56 & \cellcolor{deepgreen!79!white}0.70 \\
        & Aesthetic & \cellcolor{deepgreen!34!white}3.34 & \cellcolor{deepgreen!89!white}3.67 & \cellcolor{deepgreen!28!white}3.30 & \cellcolor{deepgreen!89!white}3.67 & \cellcolor{deepgreen!0!white}3.13 & \cellcolor{deepgreen!64!white}3.52 & \cellcolor{deepgreen!57!white}3.48 & \cellcolor{deepgreen!100!white}3.74 \\
        \hline
        \multirow{3}{*}{LeX-Bench\textsubscript{medium}} & PNED~$\downarrow$ & \cellcolor{deepgreen!26!white}5.30 & \cellcolor{deepgreen!85!white}3.87 & \cellcolor{deepgreen!12!white}5.53 & \cellcolor{deepgreen!79!white}4.03 & \cellcolor{deepgreen!0!white}5.52 & \cellcolor{deepgreen!26!white}5.10 & \cellcolor{deepgreen!29!white}4.85 & \cellcolor{deepgreen!53!white}4.43\\
        & Recall~$\uparrow$ & \cellcolor{deepgreen!50!white}0.39 & \cellcolor{deepgreen!100!white}0.52 & \cellcolor{deepgreen!40!white}0.35 & \cellcolor{deepgreen!100!white}0.52 & \cellcolor{deepgreen!0!white}0.19 & \cellcolor{deepgreen!43!white}0.33 & \cellcolor{deepgreen!13!white}0.22 & \cellcolor{deepgreen!43!white}0.32 \\
        & Aesthetic & \cellcolor{deepgreen!54!white}3.66 & \cellcolor{deepgreen!89!white}3.85 & \cellcolor{deepgreen!56!white}3.67 & \cellcolor{deepgreen!94!white}3.88 & \cellcolor{deepgreen!0!white}3.37 & \cellcolor{deepgreen!72!white}3.76 & \cellcolor{deepgreen!59!white}3.69 & \cellcolor{deepgreen!100!white}3.91 \\
        \hline
        \multirow{3}{*}{LeX-Bench\textsubscript{hard}} & PNED~$\downarrow$ & \cellcolor{deepgreen!0!white}13.38 & \cellcolor{deepgreen!64!white}9.49 & \cellcolor{deepgreen!3!white}13.12 & \cellcolor{deepgreen!66!white}9.42 & \cellcolor{deepgreen!21!white}11.49 & \cellcolor{deepgreen!40!white}10.66 & \cellcolor{deepgreen!40!white}10.27 & \cellcolor{deepgreen!50!white}9.87 \\
        & Recall~$\uparrow$ & \cellcolor{deepgreen!48!white}0.19 & \cellcolor{deepgreen!95!white}0.30 & \cellcolor{deepgreen!33!white}0.15 & \cellcolor{deepgreen!100!white}0.30 & \cellcolor{deepgreen!0!white}0.07 & \cellcolor{deepgreen!5!white}0.12 & \cellcolor{deepgreen!0!white}0.07 & \cellcolor{deepgreen!10!white}0.08 \\
        & Aesthetic~$\uparrow$ & \cellcolor{deepgreen!29!white}3.60 & \cellcolor{deepgreen!100!white}3.96 & \cellcolor{deepgreen!37!white}3.64 & \cellcolor{deepgreen!98!white}3.95 & \cellcolor{deepgreen!0!white}3.45 & \cellcolor{deepgreen!82!white}3.87 & \cellcolor{deepgreen!35!white}3.63 & \cellcolor{deepgreen!90!white}3.91 \\
    \bottomrule
    \end{tabular}}
    \caption{\textbf{Performance comparison of FLUX.1 [dev]~\cite{flux2024} and LeX-FLUX; Lumina-Image 2.0~\cite{lumina2} and LeX-Lumina.} Note that recall is computed based on NED thresholding at 0.3 to mitigate the impact of minor character errors. The color intensity in each cell indicates the relative performance, with darker green representing better performance (lower PNED or higher Recall/Aesthetic scores) and lighter green representing worse performance, with colors scaled between the best and worst values in each row.
}
    % \vspace{-1em}
    \label{tab:prompten}
\end{table*}

\begin{figure*}[htbp]
    \centering
    \includegraphics[width=\linewidth]{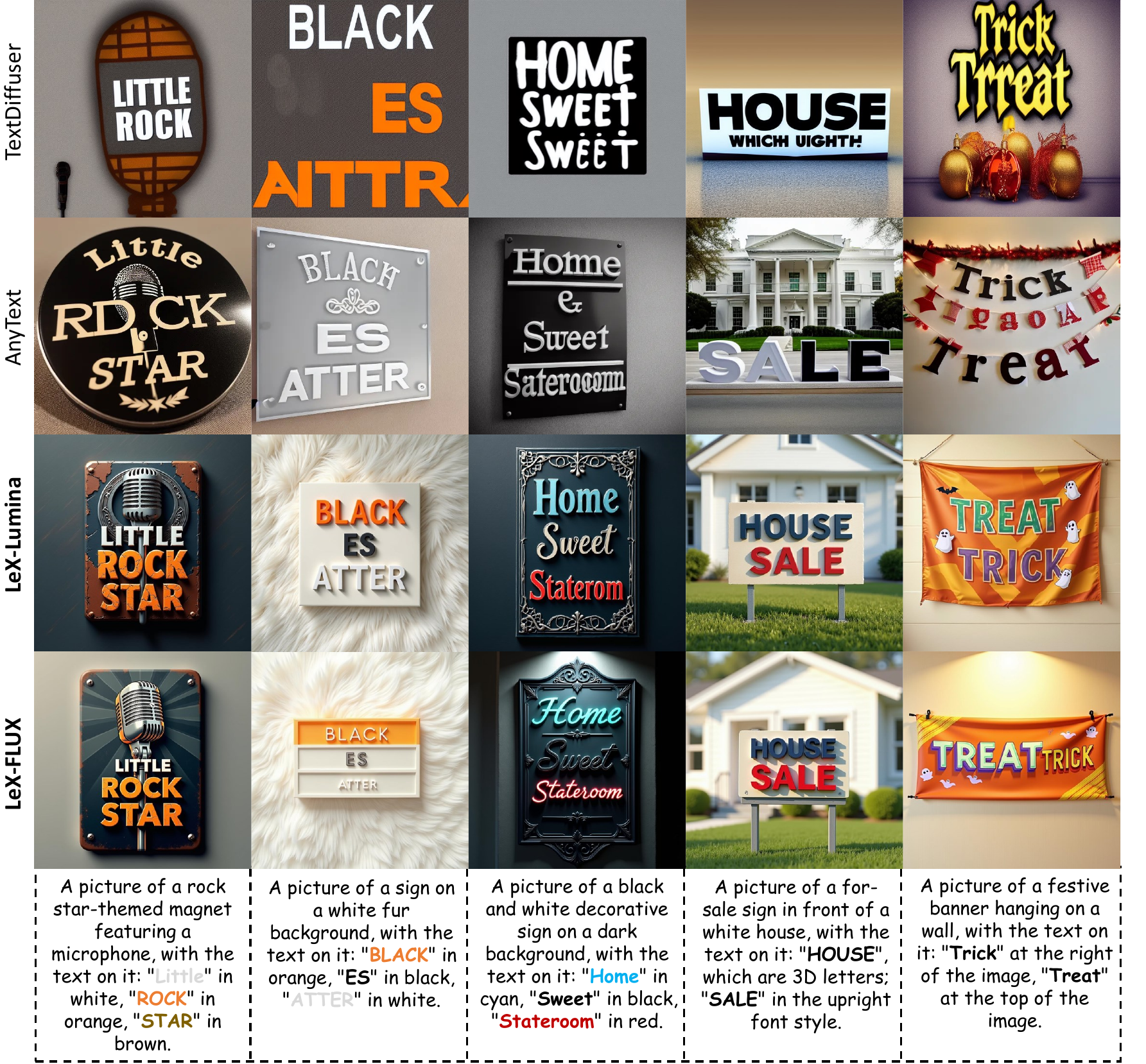}
    \caption{\textbf{Qualitative comparison between LeX-Lumina, LeX-FLUX and glyph-conditioned models.} We compare our models with AnyText~\cite{tuo2023anytext} and TextDiffuser~\cite{chen2023textdiffuser} for five different prompts. We observe that our models generally achieve high fidelity, better text attribute controllability and higher aesthetics.}
    \label{fig:com}
\end{figure*}

\begin{figure*}[htbp]
    \centering
    \includegraphics[width=\linewidth]{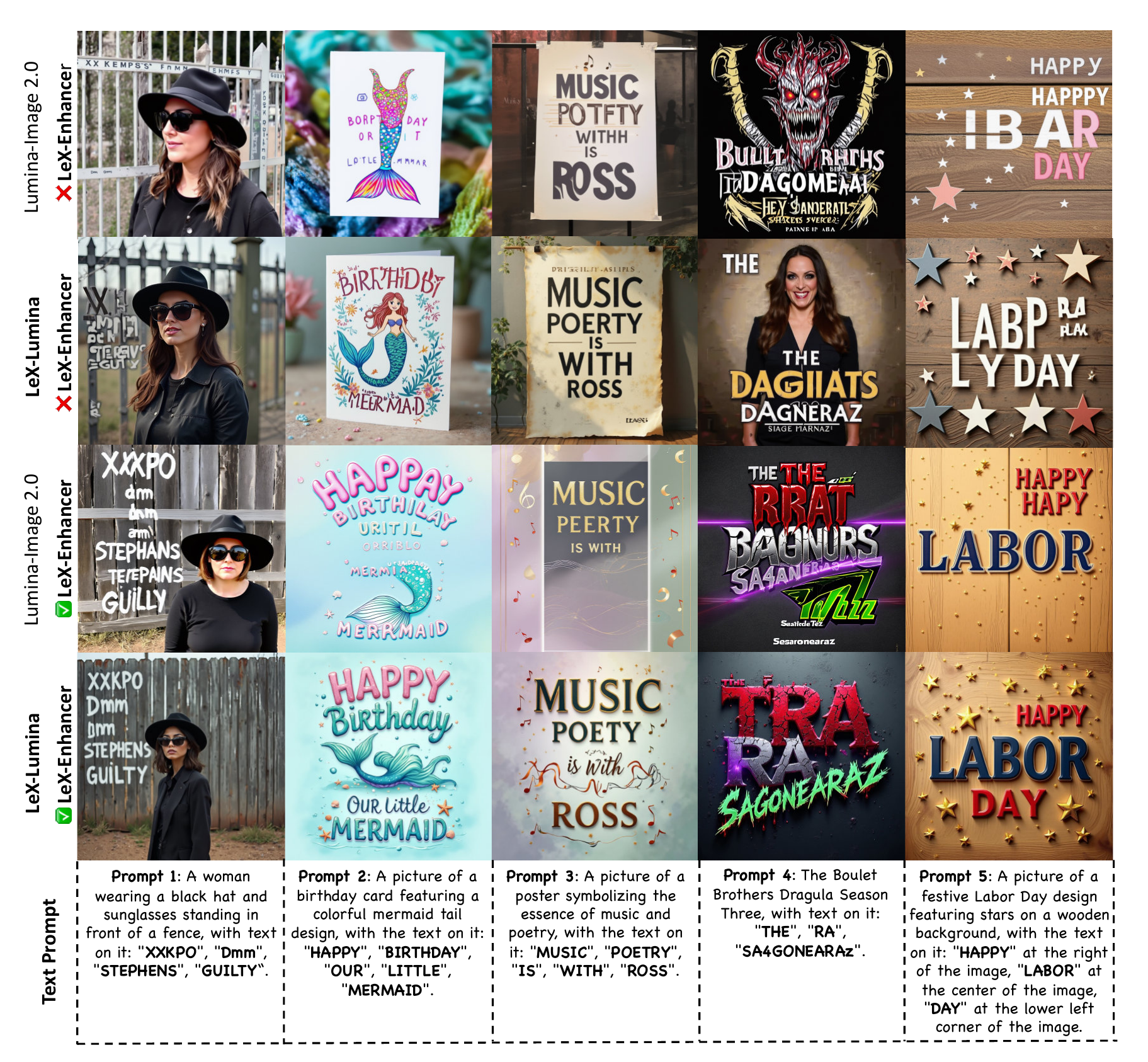}
    \caption{\textbf{Qualitative comparison between Lumina-Image 2.0~\cite{lumina2} and LeX-Lumina.} The first column shows Lumina-Image-2.0 without LeX-Enhancer using Simple Caption; the second column shows the trained LeX-Lumina without LeX-Enhancer using Simple Caption; the third column shows Lumina-Image-2.0 with LeX-Enhancer enabled; and the fourth column shows LeX-Lumina with LeX-Enhancer enabled. We observe that (1) LeX-Lumina exhibits a better text rendering capability in terms of text fidelity and aesthetics; (2) LeX-Enhancer exhibits a strong capability for enhancing simple prompts.}
    \label{fig:comp}
\end{figure*}

\begin{figure*}[htbp]
    \centering
    \includegraphics[width=0.99\linewidth]{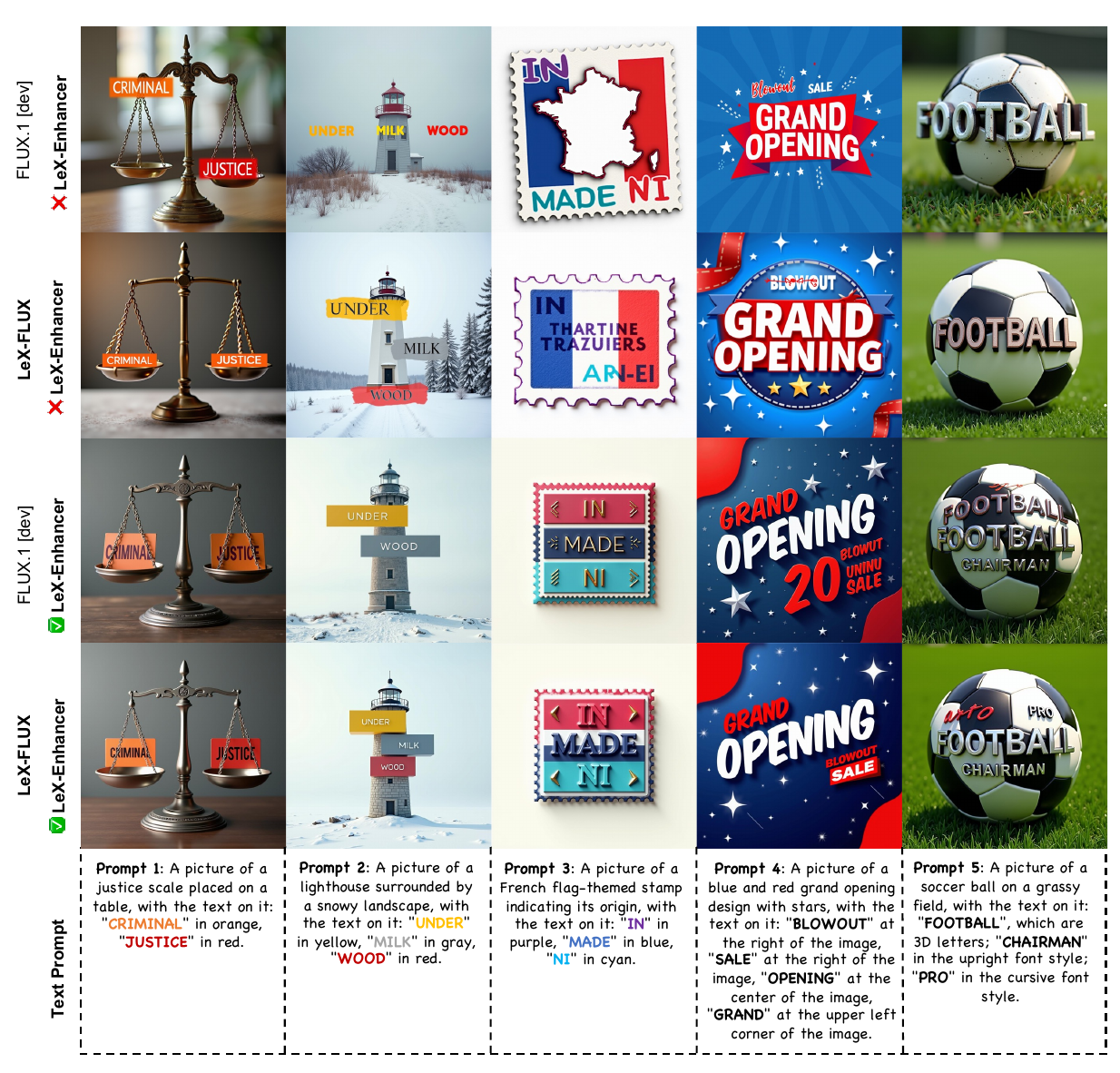}
    \caption{\textbf{Qualitative comparison between FLUX.1 [dev]~\cite{flux2024} and LeX-FLUX.} The first column shows FLUX.1 [dev] without LeX-Enhancer using Simple Caption; the second column shows the trained LeX-FLUX without LeX-Enhancer using Simple Caption; the third column shows FLUX.1 [dev] with LeX-Enhancer enabled; and the fourth column shows LeX-FLUX with LeX-Enhancer enabled. We observe that (1) LeX-FLUX exhibits a better text rendering capability in terms of text fidelity and text attributes controllability; (2) LeX-Enhancer exhibits a strong capability for enhancing simple prompts.}
    \label{fig:lexflux}
\end{figure*}

\subsection{Comparisons}
\paragraph{Comparison with state-of-the-art models.}
We compare LeX-FLUX and LeX-Lumina with glyph-conditioned models, including ControlNet~\cite{yang2023glyphcontrol}, TextDiffuser~\cite{chen2023textdiffuser}, GlyphControl~\cite{yang2023glyphcontrol}, and AnyText~\cite{tuo2023anytext}, on the AnyTextBench benchmark. As shown in \tabref{tab:compare}, glyph-conditioned methods generally achieve higher text rendering accuracy by leveraging explicit glyph information. Nevertheless, even without such glyph guidance, our models achieve competitive results; for instance, LeX-FLUX reaches an accuracy score of 0.7110, which is close to AnyText’s 0.7239. Furthermore, our methods outperform glyph-conditioned baselines in prompt-image alignment, as indicated by CLIPScore~\cite{hessel2021clipscore}—LeX-FLUX achieves a score of 0.8918, surpassing AnyText’s 0.8841.

\figref{fig:com} presents a qualitative comparison between LeX-Lumina, LeX-FLUX, and other glyph-conditioned models. Overall, LeX-Lumina and LeX-FLUX consistently demonstrate superior performance in text clarity, controllability, and visual aesthetics. They produce sharper and more accurate text, while existing methods like TextDiffuser~\cite{chen2023textdiffuser} and AnyText~\cite{tuo2023anytext} often yield blurry or distorted results. In terms of text attributes, both LeX models exhibit strong control over color, font style, and positioning—for instance, accurately rendering colored text in the first three examples where other models fail to even produce legible outputs. Furthermore, LeX-Lumina and LeX-FLUX generate more coherent and visually appealing compositions by effectively integrating text with complex backgrounds. Notably, LeX-FLUX blends text seamlessly with a 3D microphone in the first case and enhances the last three examples with refined lighting effects. These results highlight the advantages of our models in generating high-quality, well-aligned text in challenging visual contexts.

\paragraph{Improvements from fine-tuning on LeX-10K.}
We quantitatively evaluate LeX-FLUX and LeX-Lumina on LeX-Bench, with results summarized in \tabref{tab:prompten}.
LeX-FLUX, obtained by fine-tuning FLUX.1 [dev] on LeX-10K, exhibits consistent improvements in text rendering accuracy, aesthetics, and controllability of text attributes. Specifically, its PNED on LeX-Bench decreases by 0.02 compared to FLUX.1 [dev], and its average aesthetic score increases by 0.09 across three standard benchmarks. In addition, the model achieves noticeable gains in color consistency, text positioning, and font fidelity relative to the base model.
LeX-Lumina achieves even more substantial gains. Its PNED scores improve by 0.41 and 0.39 on the traditional benchmark (\eg, SimpleBench, CreateBench, and AnyText) and LeX-Bench, respectively. In terms of recall, it outperforms the baseline by 0.04 and 0.02 on the same benchmarks. Moreover, LeX-Lumina exhibits significantly enhanced controllability over text attributes compared to Lumina-Image 2.0.
Overall, while both models benefit from the proposed dataset and fine-tuning, LeX-Lumina exhibits broader and more significant improvements across all metrics.
We further compare the text rendering quality of Lumina-Image 2.0~\cite{lumina2} with LeX-Lumina (see \figref{fig:comp}), and FLUX.1 [dev]~\cite{flux2024} with LeX-FLUX (see \figref{fig:lexflux}), to assess the visual improvement. Both LeX-Lumina and LeX-FLUX produce sharper, more legible text with better alignment and fewer visual artifacts. These results clearly demonstrate the effectiveness of fine-tuning on LeX-10K in enhancing text rendering quality.

\paragraph{Human preference study.}

% In \figref{fig:human_preference}, we collected and visualized the human preference result between LeX-Lumina and Lumina-Image-2.0. The results showed that images generated by LeX-Lumina were preferred regarding text accuracy and aesthetics. 

In the experiments, we evaluated the performance of LeX-Lumina against Lumina-Image 2.0~\cite{lumina2} using aesthetic metrics and OCR-related metrics, including PNED and Recall. To further investigate whether LeX-Lumina demonstrates a clear advantage over Lumina-Image 2.0 in terms of human preference, we conducted a user study.

In this study, human annotators were presented with 40 pairs of images, where each pair consisted of outputs generated by Lumina-Image 2.0 and LeX-Lumina given the same prompt. For each pair, the annotators were asked to answer three questions:
(1) \textit{Which image has a higher aesthetic quality?}
(2) \textit{Which image renders the text more accurately?}
(3) \textit{Which image renders the text more completely with respect to the prompt?}
For each question, annotators could choose from four options: \textit{Image 1, Image 2, Both good, or Both bad}. These questions were designed to evaluate the models' capabilities in terms of aesthetics (Aesthetics), text rendering accuracy (Accuracy), and text rendering completeness (Recall).

The results of the user study are summarized in \figref{fig:human_preference}. As shown, LeX-Lumina achieved win rates of 40\%, 26\%, and 36\% for aesthetic quality, text rendering accuracy, and text rendering completeness, respectively. These results demonstrate that fine-tuning Lumina-Image 2.0 on the LeX-10K dataset produces LeX-Lumina, which achieves superior performance in human preference metrics compared to the original model.

\begin{table}[tbp]
    \centering
    \adjustbox{max width=\textwidth}{
    \begin{tabular}{c|c|cc}
    \toprule
        Data Scale & Metric & \multicolumn{2}{c}{LeX-Lumina} \\
    \midrule
    LeX-Enhancer & - & \ding{55} & \ding{51} \\
    \midrule
       \multirow{3}{*}{0}  & PNED~$\downarrow$ & 6.13 & \textbf{5.54} \\
       & Recall~$\uparrow$  & 0.23 & \textbf{0.33} \\
       & Aesthetic~$\uparrow$  & 3.29 & \textbf{3.70} \\
       \hline
       \multirow{3}{*}{1K}  & PNED~$\downarrow$ & 5.68 & \textbf{5.21} \\
       & Recall~$\uparrow$  & 0.24 & \textbf{0.33} \\
       & Aesthetic~$\uparrow$ & 3.62 & \textbf{3.86} \\
       \hline
       \multirow{3}{*}{5K}  & PNED~$\downarrow$ & 5.79 & \textbf{5.23} \\
       & Recall~$\uparrow$  & 0.25 & \textbf{0.33} \\
       & Aesthetic~$\uparrow$  & 3.45 & \textbf{3.79} \\
       \hline
       \multirow{3}{*}{10K}  & PNED~$\downarrow$ & 5.60 & \textbf{5.15} \\
       & Recall~$\uparrow$  & 0.27 & \textbf{0.35} \\
       & Aesthetic~$\uparrow$ & 3.56 & \textbf{3.83} \\
    \bottomrule
    \end{tabular}}
    \caption{\textbf{Performance comparison on LeX-Bench with different training data scales.} When scaling the training data, we observe the improving performance in terms of text rendering fidelity and aesthetics.}
    \label{tab:scaling}
    % \vspace{-2em}
\end{table}

\subsection{Ablations}

% \begin{figure*}
%     \centering
%     \includegraphics[width=\linewidth]{figures/comp.pdf}
%     \caption{\textbf{Qualitative comparison between Lumina-Image 2.0 and LeX-Lumina.} The first column shows Lumina-Image-2.0 without LeX-Enhancer using Simple Caption; the second column shows the trained LeX-Lumina without LeX-Enhancer using Simple Caption; the third column shows Lumina-Image-2.0 with LeX-Enhancer enabled; and the fourth column shows LeX-Lumina with LeX-Enhancer enabled. We observe that (1) LeX-Lumina exhibits a better text rendering capability in terms of text fidelity and aesthetics; (2) LeX-Enhancer exhibits a strong capability for enhancing simple prompts.}
%     \label{fig:comp}
% \end{figure*}

% \begin{figure*}
%     \centering
%     \includegraphics[width=0.99\linewidth]{figures/comp-flux.pdf}
%     \caption{\textbf{Qualitative comparison between FLUX.1 [dev] and LeX-FLUX.} The first column shows FLUX.1 [dev] without LeX-Enhancer using Simple Caption; the second column shows the trained LeX-FLUX without LeX-Enhancer using Simple Caption; the third column shows FLUX.1 [dev] with LeX-Enhancer enabled; and the fourth column shows LeX-FLUX with LeX-Enhancer enabled. We observe that (1) LeX-FLUX exhibits a better text rendering capability in terms of text fidelity and text attributes controllability; (2) LeX-Enhancer exhibits a strong capability for enhancing simple prompts.}
%     \label{fig:lexflux}
% \end{figure*}

% \subsection{Qualitative Analysis}

% \textcolor{cyan}{add a figure here}
\paragraph{Effectiveness of Lex-Enhancer.}

In \secref{sec:dataset}, we demonstrated that prompts enhanced using DeepSeek-R1 significantly improve the aesthetic quality and text accuracy of generated images. To further evaluate the prompt enhancement capability of our trained LeX-Enhancer, we conducted a series of experiments across multiple benchmarks.
Specifically, we evaluated prompts enhanced by LeX-Enhancer on three benchmarks—SimpleBench~\cite{yang2023glyphcontrol}, AnyText-Benchmark~\cite{tuo2023anytext}, and LeX-Bench—using four models: FLUX.1 [dev], LeX-FLUX, Lumina-Image 2.0, and LeX-Lumina, both with and without LeX-Enhancer. Detailed results are provided in \tabref{tab:prompten}.
Across all four models and six benchmarks, prompts enhanced by LeX-Enhancer consistently led to significant improvements in text rendering accuracy, aesthetic quality, and controllability of text attributes. Specifically, FLUX.1 [dev] achieved a 1.43-point gain in PNED on LeX-Bench-medium, LeX-FLUX saw a 19.37\% increase in text style scores on LeX-Bench, Lumina-Image 2.0 obtained a 20\% improvement in Recall on CreateBench, and LeX-Lumina achieved a 9.64\% boost in text position scores on LeX-Bench.
In addition, comparisons with glyph-conditioned models reveal that LeX-Enhancer significantly improves both the Acc. scores~\cite{tuo2023anytext} and CLIPScore~\cite{hessel2021clipscore} of LeX-FLUX and LeX-Lumina, as shown in \tabref{tab:compare}. 
% Furthermore, for models trained on synthetic datasets of varying scales, we observed that LeX-Enhancer consistently enhances the models' capabilities in text rendering accuracy and aesthetic quality (see \tabref{tab:scaling}).
\figref{fig:comp} and \figref{fig:lexflux} spresent qualitative comparisons of images generated by Lumina-Image 2.0 and LeX-Lumina, as well as FLUX.1 [dev] and LeX-FLUX, with and without LeX-Enhancer, further illustrating the effectiveness of our method.
\textit{These results underscore the critical role of prompt quality in visual text generation and highlight the value of prompt enhancement for training effective text-to-image models.}

\paragraph{Study on training data scale}
% 1k, 5k, 10k
As shown in \tabref{tab:scaling}, we conduct an ablation study on Lumina-Image 2 by training it with 1K, 5K, and 10K text-image pairs to evaluate the impact of data scale on text rendering quality. We observe clear improvements in text accuracy, layout alignment, and aesthetic quality as the size of the dataset increases. Smaller datasets tend to cause more distortions and inconsistencies, while larger datasets lead to better fidelity and spatial consistency.
Based on these findings, we fine-tune our models on the larger 10K dataset. In addition to scaling data, we find that LeX-Enhancer consistently improves performance when integrated with LeX-Lumina. Furthermore, with the proposed scalable text-image data construction method, we expect to obtain more high-quality samples to further boost model performance.
\section{Conclusion}
% We developed R1-enhanced prompt augmentation, a high-quality text-image dataset, and a scalable training pipeline to improve text rendering in text-to-image models. By leveraging R1-enhanced prompts and post-aligned text-image pairs, we curated a high-fidelity dataset that significantly enhances text generation performance. Our experiments show that scaling high-quality training data leads to consistent improvements in text accuracy, layout alignment, and aesthetic quality. Additionally, proposed a comprehensive benchmark to evaluate models' text rendering ability regarding accuracy, aesthetic, and text attributes controllability. The effectiveness of our approach highlights the crucial role of high-quality data and enhanced prompting in advancing text rendering capabilities in text-to-image generation.

We present LeX-Art, a data-centric framework that systematically improves the text rendering capability of text-to-image models. By leveraging DeepSeek-R1, we construct LeX-10K, a high-quality dataset of 10K refined text-image pairs through enriched prompts and multi-stage filtering.
Building on this, we develop LeX-Enhancer, a lightweight prompt enrichment model trained on 60K enhanced prompts, and fine-tune two models—LeX-FLUX and LeX-Lumina—to achieve strong performance in text fidelity, layout, and aesthetics.
To evaluate visual text generation, we introduce LeX-Bench, a benchmark covering fidelity, aesthetics, and attribute control, along with PNED, a robust metric for OCR-based text accuracy.
Extensive results show that LeX-Art offers a scalable and effective solution for high-quality visual text generation, bridging the gap between prompt expressiveness and rendering precision.
%
%In this work, we enhance the text rendering capability of text-to-image foundation models from a data-centric perspective. Specifically, we leverage DeepSeek-R1 to synthesize a set of domain-specific, highly detailed image descriptions. Subsequently, we design a series of post-processing modules, resulting in LeX-10K, a synthetic dataset comprising 10,000 high-quality text-image samples. Using the prompt-enhancing data generated by DeepSeek-R1, we fine-tune a locally deployable prompt enhancer, LeX-Enhancer. Furthermore, leveraging LeX-10K, we fine-tune two existing models, FLUX.1 [dev] and Lumina-Image 2.0, to obtain LeX-FLUX and LeX-Lumina, respectively.

%To rigorously evaluate text rendering performance, we construct a comprehensive benchmark, LeX-Bench, which assesses text rendering accuracy, aesthetics, and controllability over text attributes. In addition, we introduce a flexible and general metric, Pairwise Normalized Edit Distance (PNED), specifically designed for evaluating text rendering precision. Extensive experiments conducted on LeX-Bench demonstrate that LeX-FLUX and LeX-Lumina achieve significant improvements across all aspects of text rendering.
{
    \small
    \bibliographystyle{ieeenat_fullname}
    \bibliography{main}
}
\newpage
% \newpage
\clearpage
\appendix
\noindent
{\large \textbf{Appendix}}
% \section{Seed Caption Curation Pipeline}

\section{Instruction Used in Prompt Enhancer}
\label{appendix:instruction}

To enhance short captions into those suitable for generating high-quality images, it is crucial to first understand the characteristics of an effective caption. Based on insights from the community~\cite{runway}, a good caption should possess the following attributes: it should be detailed, well-structured, rich in knowledge, and written in a descriptive tone. To achieve this, we devised the instruction shown below to transform concise captions into more comprehensive and informative ones.

\begin{user_example}[frametitle={Prompt Enhancer Template}]
{
    \fontsize{10pt}{9pt}\selectfont \textbf{Simple Caption:} \textless Simple Caption\textgreater\\
    \\Above is the simple caption of an image with text. Please deduce the detailed description of the image based on this simple caption. \textbf{Note}: \\
    % {\small
    \textbf{1}.The description should only include visual elements and should not contain any extended meanings. \\
    \textbf{2}.The visual elements should be as rich as possible, such as the main objects in the image, their respective attributes, the spatial relationships between the objects, lighting and shadows, color style, any text in the image and its style, etc. \\
    \textbf{3}.The output description should be a single paragraph and should not be structured. \\
    \textbf{4}.The description should avoid certain situations, such as pure white or black backgrounds, blurry text, excessive rendering of text, or harsh visual styles. \\
    \textbf{5}.The detailed caption should be human-readable and fluent. \\
    \textbf{6}.Avoid using vague expressions such as ``may be" or ``might be"; the generated caption must be in a definitive, narrative tone. \\
    \textbf{7}.Do not use negative sentence structures, such as ``there is nothing in the image," etc. The entire caption should directly describe the content of the image. \\
    \textbf{8}.The entire output should be limited to 200 words.
    % }
}
\end{user_example}

\section{Instruction Used in Knowledge-Augmented Recaption}

As mentioned in the paper, the captions enhanced by DeepSeek-R1~\cite{guo2025deepseek} contain rich knowledge, such as font styles, layout configurations, color schemes, and more. However, due to the inherent limitations of FLUX.1 [dev]~\cite{flux2024}, the images generated using these prompts do not always align perfectly with the prompts themselves. To address this issue, we designed a knowledge-augmented recaptioning step to further refine the prompts, ensuring they are fully aligned with the image content while preserving the rich knowledge introduced by DeepSeek-R1~\cite{guo2025deepseek}.

\label{appendix:postalign}
\begin{user_example}[frametitle={Knowledge-Augmented Recaption Template}]
{
\fontsize{10pt}{9pt}\selectfont \textbf{Image:} \textless Image\textgreater\\
\fontsize{10pt}{9pt}\selectfont \textbf{Original Caption:} \textless Original Caption\textgreater\\
\fontsize{10pt}{9pt}\selectfont \textbf{Instruction:}
This is the caption of an image with text rendered on and the corresponding image. There might be some artifacts in the image. For example, some of the texts were not be rendered correctly in the generated image. I need you to refer to the provided caption and corresponding generated image, refine the caption based on the generated image. \fontsize{10pt}{9pt}\selectfont \textbf{Note:} \textbf{1}.The refined caption should fully describe the generated image. \\
\textbf{2}.In the refinded caption, the misalignment of the original caption and the generated image should be fixed and the other visual details should be keeped. \\
\textbf{3}.Directly output the refined caption. \\
\textbf{4}.The output description should be a single paragraph and should not be structured. \\
\textbf{5}.The entire output should be limited to 200 words.
}
\end{user_example}

\section{Instruction Used in Prompt Refinement (LeX-Bench Curation)}
\label{appendix:lexbench}

Since Florence-2~\cite{xiao2024florence} may include textual content from the image in its descriptions or occasionally produce outputs that lack fluency, we leveraged GPT-4o~\cite{hurst2024gpt} to refine and adjust the image descriptions generated by Florence-2.

\begin{user_example}[frametitle={Prompt Refinement Template}]
{
Below is the caption of an image along with the text I provided. Please revise this caption, ensuring that the revised caption does not include the text I provided while maintaining the original meaning as much as possible. \textbf{Note}: \\
\textbf{1}.The refined caption should be kept brief and concise, and it should describe an image containing no text. \\
\textbf{2}.Directly give me the refined caption. \\
\textbf{3}.Maybe the refined caption could start with ``An image of..." or ``A picture of...". \\
\textbf{4}.Remember the provided text must not be included in the refined caption. \\
\textbf{5}.The refined caption should be fluent. \\
\textbf{6}.Most importantly: the refined caption must not contain any text to be rendered on the image. \\\\
\fontsize{10pt}{9pt}\selectfont \textbf{Simple Caption:} \textless Simple Caption \textgreater\\
\\
\fontsize{10pt}{9pt}\selectfont \textbf{Text:} \textless OCR Results \textgreater\\

}
\end{user_example}

\section{The Format of Text Conditions}
\label{appendix:text_conditions}
We design three conditions for text presentation, focusing on \textbf{color}, \textbf{font}, and \textbf{position}. This section provides a detailed description of these three conditions. It is important to note that the evaluation of color and font is conducted using GPT-4o, and a series of tests are performed to ensure that GPT-4o can generate accurate and reliable results. Below, we elaborate on the specific design of each condition.

\textbf{Color.} For the color condition, we select a set of 12 colors as condition. The color condition and the format of caption are as follows:
\begin{user_example}[frametitle={Color Condition List}]
{
colors = [``red", ``blue", ``green", ``yellow", ``orange", ``purple", ``pink", ``brown", ``black", ``white", ``gray", ``cyan"].
}
\end{user_example}

\begin{user_example}[frametitle={Prompt Format with Color Condition}]
{
\textit{\{Image Caption\}, with the text on it: \{Text Caption\} in \{color\}; \{Text Caption\}in \{color\} ...}. 
}
\end{user_example}

This allows for a systematic exploration of how different colors interact with fonts and other design elements.

\textbf{Font.} In the font condition, we employ five pairs of contrasting font styles to ensure diversity and clarity. To maintain design consistency, opposing font styles from the same pair are not presented within the same image. For instance, ``3D style" and ``flat style" are never displayed together, as their will complicate the visual coherence of the image. These pairs and format are as follows:

\begin{user_example}[frametitle={Font Condition List}]
{
fonts = [

[``cursive style", ``block style"], 

[``3D style", ``flat style"], 

[``sans-serif", ``serif"], 

[``upright", ``slant"],  

[``rounded", ``angular"]

]
}
\end{user_example}

\begin{user_example}[frametitle={Prompt Format with Font Condition}]
{
\textit{\{Image Caption\}, with the text on it: \{Text Caption\}, \{font description\}; \{Text Caption\}, \{font description\}; ...}.

The detailed \textit{\{font description\}} are as follows:

``cursive style": in the cursive font style

``block style": in the block font style

``3D style": which are 3D letters

``flat style": which sits flat

``sans-serif": in the sans-serif style

``serif": in the serif style

``upright": in the upright font style

``slant": in the slant font style

``rounded": in the rounded font style

``angular": in the angular font style
}
\end{user_example}

\textbf{Position.} Finally, for the position condition, we select a variety of spatial placements for the text. These conditions and format are as follows:

\begin{user_example}[frametitle={Position Condition List}]
{
positions = [``top", ``bottom", ``left", ``right", ``upper left corner", ``lower left corner", ``upper right corner", ``lower right corner", ``center"].
}
\end{user_example}

\begin{user_example}[frametitle={Prompt Format with Position Condition}]
{  
\textit{\{Image Caption\}, with the text on it: \{Text Caption\}, at the \{position\} of the image;  \{Text Caption\}, at the \{position\} of the image...}.
}
\end{user_example}

This systematic approach ensures that the text placement is both diverse and well-documented for further analysis.

% \section{Preference Evaluation Design}
% \label{appendix:preference_eval}

\section{Normalized Edit Distance Algorithm}
\label{appendix:NED}
Edit Distance (ED), also known as Levenshtein Distance, measures the minimum number of operations required to transform one string into another. The allowed operations include character insertions, deletions, and substitutions, each contributing a unit cost to the total edit distance. This metric is widely used for text similarity evaluation, particularly in tasks such as optical character recognition (OCR), spelling correction, and machine translation.

To improve interpretability and enable comparisons across different string lengths, Normalized Edit Distance (NED) is introduced. NED normalizes the computed edit distance by dividing it by the maximum length of the two strings, ensuring a value between 0 and 1, where 0 indicates identical strings and 1 represents completely different strings. The formal computation of NED is shown in Algorithm \ref{alg:ned}, where a two dimensional dynamic programming (DP) table is used to iteratively compute the minimum edit cost between two input strings.

\begin{algorithm}
\caption{Normalized Edit Distance (NED)}
\label{alg:ned}
\begin{algorithmic}[1]
\State Given two strings $a$ and $b$ of lengths $n$ and $m$. Initialize DP table $D \in \mathbb{R}^{(n+1) \times (m+1)}$. Define edit cost $c$. Initialize $D_{i,0} \gets i$, $D_{0,j} \gets j$ for all $i, j$.
\For{$i = 1$ to $n$}
    \For{$j = 1$ to $m$}
        \State $D_{i,j} \gets \min(D_{i-1,j} + 1, D_{i,j-1} + 1, D_{i-1,j-1} + I(a_i \neq b_j))$
    \EndFor
\EndFor
\State \Return $D_{n,m} / \max(n, m)$
\end{algorithmic}
\end{algorithm}

\section{Data Samples from LeX-10K}
\label{appendix:lex10k}
\begin{figure*}
    \centering
    \includegraphics[width=\linewidth]{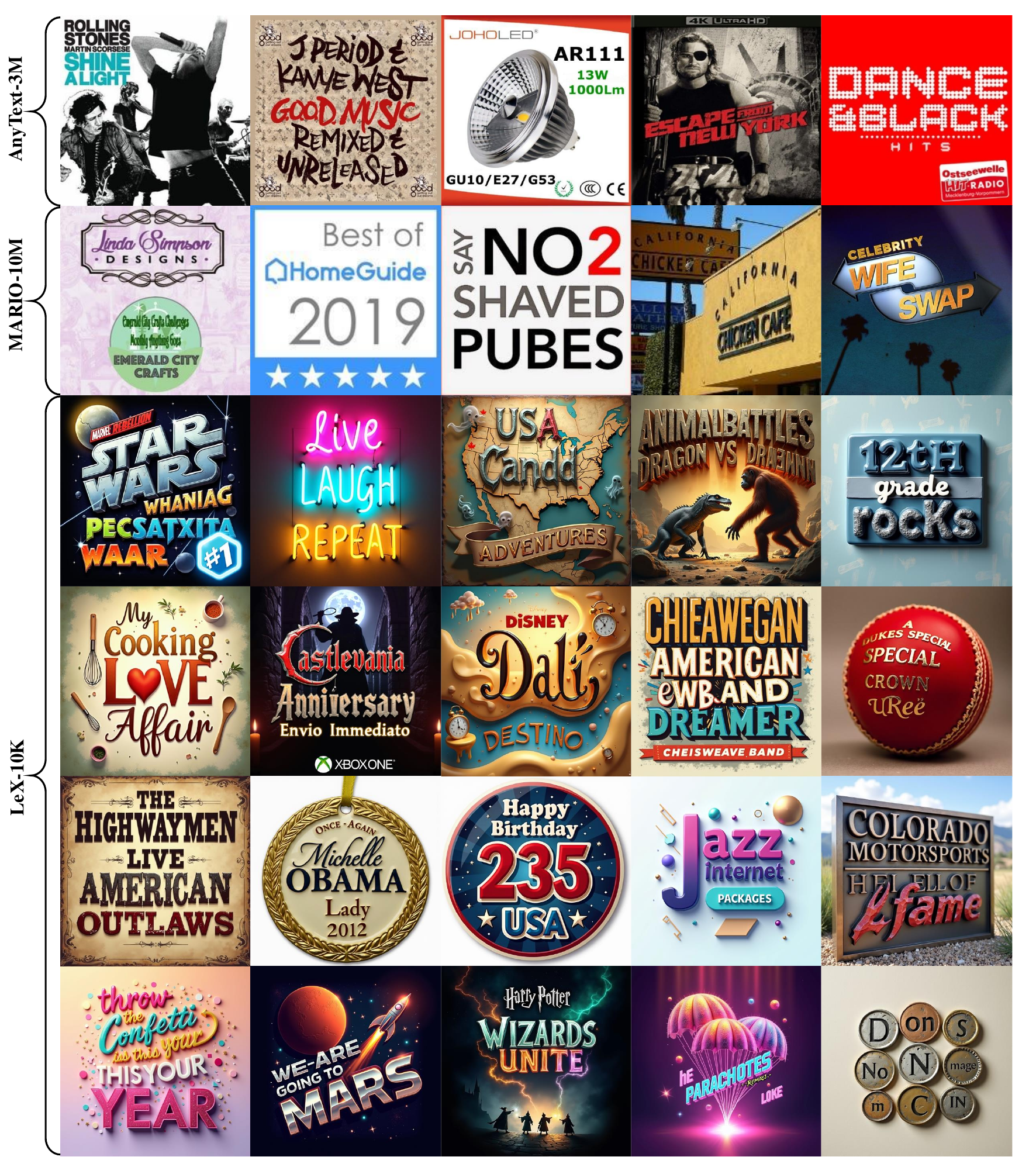}
    \caption{Comparison of data samples from AnyWord-3M~\cite{tuo2023anytext}, MAION-10M~\cite{chen2023textdiffuser} and LeX-10K.}
    \label{fig:lex10k}
    \vspace{3em}
\end{figure*}

In \figref{fig:lex10k}, we show the comparison of data samples from AnyWord-3M~\cite{tuo2023anytext}, MARIO-10M~\cite{chen2023textdiffuser}, and LeX-10K. It is obvious that images from LeX-10K are better, regarding aesthetics and diversity.

\section{The Evaluation of Text Attributes}
\label{appendix: condition_eval}
The evaluation of color, font, and position conditions is conducted using OCR and VQA. Specifically, OCR is first employed to detect all textual elements within the image. The detected words are then split by spaces and subjected to fuzzy matching against the ground truth (GT) texts provided in the prompt. After matching, it is checked whether the GT condition for each word corresponds to how the text is actually presented in the image.

\textbf{Color Score.} Once the detected words are matched with the GT, the bounding box of the matched text is cropped from the image. To ensure accuracy, the cropped area is slightly larger than the detected bounding box. The cropped region is then passed to GPT-4o for verification using a structured query format:

\begin{user_example}[frametitle={VQA Format of Color Attribute}]
{
\textit{The text ``\{text\}" is in the color of \{color\}? Answer me using ``yes" or ``no".}
}
\end{user_example}

Here, \textit{``{text}"} represents the word detected by OCR, and \textit{``{color}"} corresponds to the color condition specified in the prompt.

\textbf{Font Score.} The matched text is cropped and passed to GPT-4o similarly. However, unlike the color evaluation, the query format is designed to accommodate font pair comparisons:

\begin{user_example}[frametitle={VQA Format of Font Attribute}]
{
\textit{The text ``\{text\}" is \{font\_A\} or \{font\_B\}? Answer me using either ``\{font\_A\}" or ``\{font\_B\}" only.}
}
\end{user_example}

Here, \textit{``\{text\}"} is the OCR-detected word, \textit{``\{font\_A\}"} represents the font specified in the prompt as the GT condition, and \textit{``\{font\_B\}"} is the corresponding paired font associated with \textit{``{font\_A}"}.

\textbf{Position Score.} The coordinate ranges corresponding to the nine predefined position conditions are set in advance. The bounding box of the OCR-detected text is compared against the coordinate range corresponding to the matched GT condition. If the bounding box falls within the specified range, the condition is considered satisfied.

\section{More Generated Samples of LeX-FLUX and LeX-Lumina}
\begin{figure*}
    \centering
    \includegraphics[width=\linewidth]{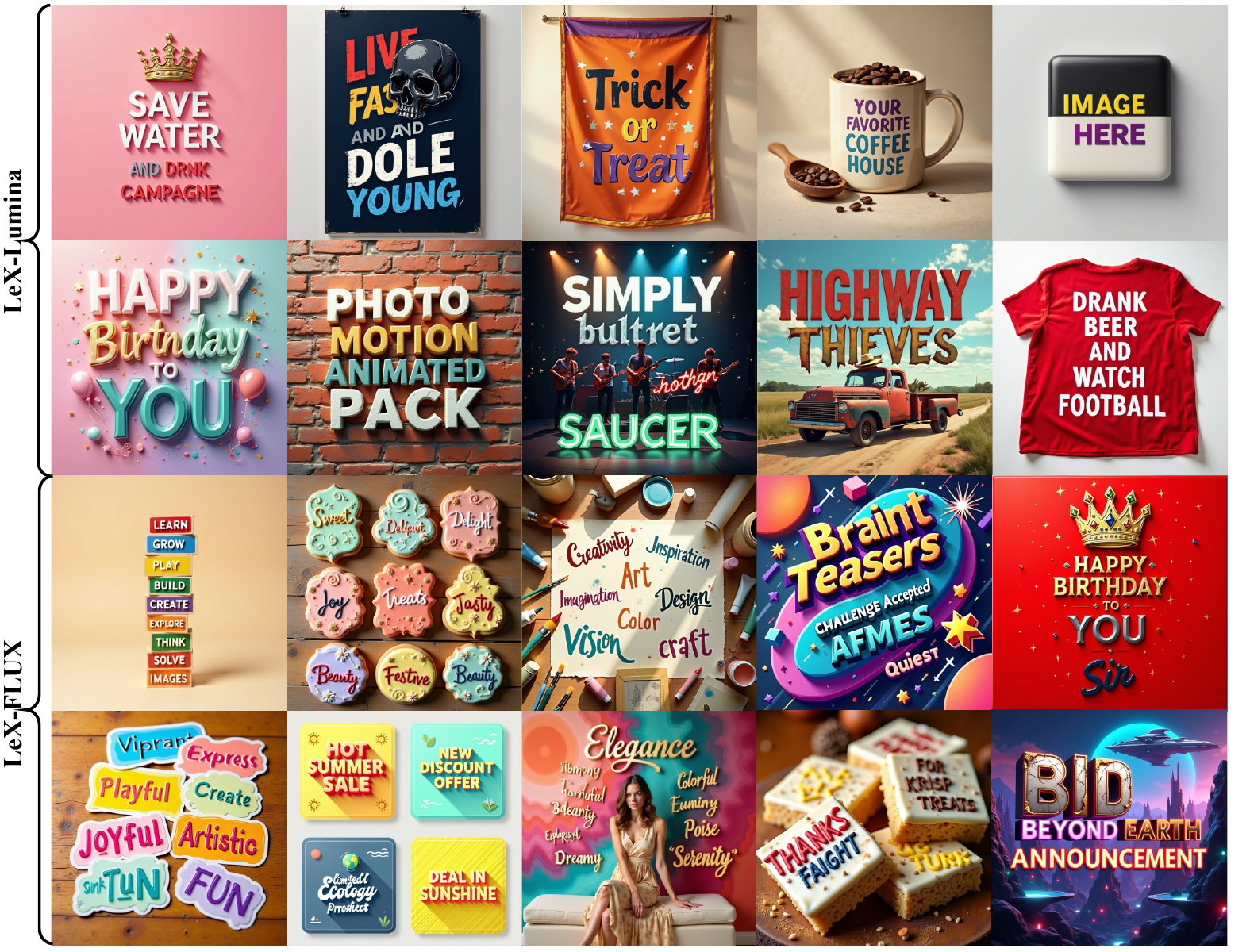}
    \caption{Showcase of text rendering results from LeX-Lumina (first two rows) and LeX-FLUX (last two rows) on text-to-image tasks. The examples demonstrate the models’ ability to generate clear, well-aligned, and aesthetically pleasing text within images. }
    \label{fig:showcase}
\end{figure*}

In \figref{fig:showcase}, we show more generated images of LeX-FLUX and LeX-Lumina.

\section{Justification of Pairwise Normalized Edit Distance (PNED) as a Metric}

To validate the effectiveness of PNED as a reliable metric for evaluating text rendering performance, we conduct a systematic empirical study through controlled perturbation experiments. Our objective is to demonstrate that PNED effectively captures the degree of deviation between detected text sequences and ground truth, while being robust to sequence order variations.

\subsection{Experimental Design}

We design a controlled experiment where we can precisely manipulate the degree of text perturbation and analyze how PNED responds to these changes. The experiment consists of the following key components:

1) \textbf{Dataset Generation}: We synthesize a dataset of $N=100$ samples, where each sample contains a list of strings with lengths randomly distributed between 1 and 20. Each string comprises lowercase ASCII characters with lengths randomly distributed between 3 and 8 characters.

2) \textbf{Perturbation Mechanism}: We implement a stochastic perturbation algorithm that accepts a parameter $\alpha \in [0,1]$ controlling the perturbation intensity. For each string, the algorithm randomly applies one of six operations with probability $\alpha$:
   - Character insertion
   - Character deletion
   - Character replacement
   - String splitting into two substrings
   - String addition (adding a new random string)
   - String deletion (removing an existing string)

3) \textbf{Evaluation Protocol}: We treat the original dataset as ground truth and the perturbed dataset as OCR predictions. We compute PNED scores between corresponding pairs under two conditions:
   - Ordered: maintaining the original sequence order
   - Shuffled: randomly permuting the sequence order to simulate OCR outputs with uncertain ordering

\subsection{Results and Analysis}

\begin{figure}[thbp]
\vspace{-1em}
    \centering
    \includegraphics[width=0.7\linewidth]{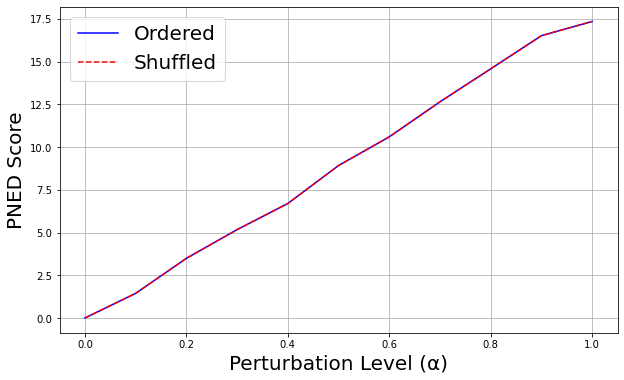}
    \caption{Validation of PNED metric through controlled perturbation experiments. The solid blue line represents PNED scores with maintained sequence order, while the dashed red line shows scores with shuffled sequences.}
    \label{fig:pned_validation}
    \vspace{-1em}
\end{figure}

\figref{fig:pned_validation} illustrates the relationship between the perturbation intensity $\alpha$ and the resulting PNED scores. Several key observations support the validity of PNED as a metric:

1) \textbf{Monotonicity}: PNED scores demonstrate a consistent monotonic increase with respect to $\alpha$, confirming that the metric effectively captures the degree of text degradation.

2) \textbf{Order Invariance}: The minimal difference between ordered and shuffled conditions (solid blue vs. dashed red lines) validates that PNED, through its optimal matching mechanism, successfully handles sequence order variations - a crucial property for OCR evaluation.

3) \textbf{Unbounded Range}: Unlike normalized metrics, PNED values are not constrained to [0,1]. Instead, they represent the raw distance between sequences, making it suitable for scenarios where absolute deviation is more meaningful than relative similarity.

4) \textbf{Sensitivity}: The metric shows appropriate sensitivity to perturbations, with distinguishable changes in PNED scores across different $\alpha$ values, indicating its effectiveness in discriminating varying degrees of OCR errors.

These empirical results demonstrate that PNED satisfies key desirable properties of an evaluation metric: monotonicity, order invariance, unboundedness, and appropriate sensitivity. Furthermore, its ability to handle variable-length sequences and account for unmatched elements makes it particularly suitable for real-world OCR evaluation scenarios where predictions may contain splits, merges, or missing characters.

% \begin{figure}[t]
% \centering
% [Figure showing PNED scores vs. perturbation level]
% \caption{Validation of PNED metric through controlled perturbation experiments. The solid blue line represents PNED scores with maintained sequence order, while the dashed red line shows scores with shuffled sequences.}
% \label{fig:pned_validation}
% \end{figure}

\section{Self-Enhancement and Knowledge Distillation}
We observe a self-enhancement effect in FLUX.1 [dev] when fine-tuned on high-quality self-distilled data. While the direct performance gain from self-training is subtle, the impact is significantly amplified when the data is used for knowledge distillation. Specifically, using FLUX.1 [dev] as a teacher model for Lumina-Image 2.0 leads to a substantial improvement in the student model’s text rendering accuracy, layout coherence, and aesthetic quality. Meanwhile, the teacher itself, FLUX.1 [dev] can be self-enhanced by the self-distilled data.

This observation highlights the scalability of our approach: as the teacher model continues to improve through iterative refinement on better-curated data, the performance gains cascade down to student models more effectively. This suggests that even small advancements in a strong base model can translate into major enhancements for smaller, more efficient models, making this distillation process a promising direction for scaling text-to-image models with improved text generation capabilities.
% \begin{figure}[t]
% \centering
% [Figure showing PNED scores vs. perturbation level]
% \caption{Validation of PNED metric through controlled perturbation experiments. The solid blue line represents PNED scores with maintained sequence order, while the dashed red line shows scores with shuffled sequences.}
% \label{fig:pned_validation}
% \end{figure}

\end{document}